\documentclass[journal]{IEEEtran}

\usepackage[T1]{fontenc}
\usepackage{graphicx}
\usepackage{url}
\usepackage{amsmath,amssymb,amsfonts}
\usepackage{multirow}
\usepackage{booktabs}
\usepackage{array}
\makeatletter\newcommand\captionof[1]{\def\@captype{#1}\caption}\makeatother
\usepackage[table]{xcolor}

\definecolor{tabfirst}{rgb}{1.0,0.70,0.70}
\definecolor{tabsecond}{rgb}{1.0,0.85,0.70}
\definecolor{tabthird}{rgb}{1.0,0.95,0.70}
\newcommand{\cellbest}[1]{\cellcolor{tabfirst}#1}
\newcommand{\cellsecond}[1]{\cellcolor{tabsecond}#1}
\newcommand{\cellthird}[1]{\cellcolor{tabthird}#1}
\newcommand{\rankswatch}[2]{{\setlength{\fboxsep}{0.4pt}\colorbox{#1}{\hspace{0.6pt}\vphantom{Ag}#2\hspace{0.6pt}}}}
\newcommand{\rankkey}{The \rankswatch{tabfirst}{best}, \rankswatch{tabsecond}{second best}, and \rankswatch{tabthird}{third best} entries are highlighted}

\definecolor{cvprblue}{rgb}{0.21,0.49,0.74}
\usepackage[colorlinks=true,linkcolor=cvprblue,citecolor=cvprblue,urlcolor=cvprblue]{hyperref}
\usepackage{orcidlink}

\usepackage[sort,nocompress]{cite}

\newcommand{\dsone}{DS0001}
\newcommand{\mevidvl}{VL-MEVID}

\begin{document}

\title{TransFIRA+: Transfer Learning for\\Biometric Recognizability Assessment}

\author{Allen~Tu\,\orcidlink{0009-0002-6532-2863},
        Kartik~Narayan\,\orcidlink{0000-0002-3095-9752},
        Joshua~Gleason\,\orcidlink{0009-0007-4162-5655},
        Jennifer~Xu\,\orcidlink{0000-0002-7374-6466},
        \\Matthew~Meyn\,\orcidlink{0000-0002-2321-1105},
        Tom~Goldstein\,\orcidlink{0000-0003-1660-9307},
        and~Vishal~M.~Patel\,\orcidlink{0000-0002-5239-692X}%
\thanks{This work has been submitted to the IEEE for possible publication. Copyright may be transferred without notice, after which this version may no longer be accessible.}%
\thanks{This work was supported by the Office of the Director of National Intelligence (ODNI), Intelligence Advanced Research Projects Activity (IARPA), via IARPA R\&D Contracts No.~2022-21102100005 and No.~56000026C0026. An earlier version of this work was presented at the 20th IEEE International Conference on Automatic Face and Gesture Recognition (FG 2026)~\cite{Tu2026TransFIRA}. (Corresponding author: Allen Tu.)}%
\thanks{Allen Tu and Tom Goldstein are with the Department of Computer Science, University of Maryland, College Park, MD, USA.}%
\thanks{Kartik Narayan and Vishal M. Patel are with the Department of Electrical and Computer Engineering, Johns Hopkins University, Baltimore, MD, USA.}%
\thanks{Jennifer Xu and Joshua Gleason are with Systems and Technology Research, Arlington, VA, USA. Part of this work was performed while Allen Tu and Matthew Meyn were with Systems and Technology Research.}}

\markboth{}{}

\makeatletter
\let\TF@oldmaketitle\@maketitle
\renewcommand{\@maketitle}{\TF@oldmaketitle\vspace{1mm}\setcounter{figure}{0}\begin{center}\expandafter\def\csname @captype\endcsname{figure}
  \includegraphics[width=\textwidth]{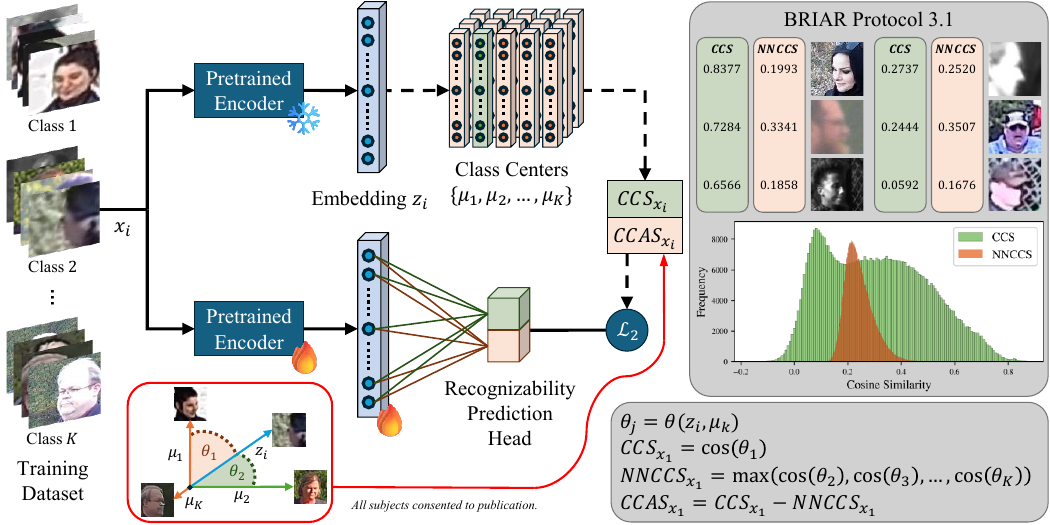}
  \caption{Visual quality does not reliably indicate whether an image is recognizable to a biometric encoder; for instance, blurred faces may remain discriminative while clear ones may fail. \textbf{TransFIRA} instead derives recognizability labels directly from a pretrained encoder's class-center geometry and finetunes the encoder to predict them. These encoder-specific scores enable $CCS$-weighted template aggregation and principled filtering at the natural cutoff $CCAS>0$, yielding state-of-the-art face verification on the long-range surveillance benchmark BRIAR Protocol~3.1~\cite{cornett2023expanding}. The same scores also enable encoder-grounded analysis of image recognizability. \textbf{TransFIRA+} extends this framework to multimodal cross-camera person identification in video, demonstrating that encoder-grounded recognizability generalizes across biometric modalities and recognition settings. Project page: \url{https://transfira.github.io/plus}.}
  \label{fig:teaser}
\end{center}\vspace{-6mm}}
\makeatother

\maketitle

\begin{abstract}
Biometric recognition in surveillance, video, and web imagery must contend with extreme variation in pose, blur, illumination, occlusion, and distance, where visual quality metrics fail to predict whether inputs are recognizable to the deployed encoder. Existing face image quality assessment methods rely on visual heuristics, curated annotations, or costly generative pipelines, leaving their predictions detached from the encoder's decision geometry. We introduce TransFIRA, a lightweight framework that grounds recognizability directly in embedding space. TransFIRA derives class-center similarity (CCS) and class-center angular separation (CCAS) from a pretrained encoder's embeddings and learns to predict them from images without quality annotations. Using predicted CCS for template weighting and CCAS>0 as a natural filtering threshold yields state-of-the-art face verification performance on BRIAR, raising TAR at 10\textsuperscript{-6} FMR from 0.085 with uniform averaging to 0.598. The predicted scores also reveal how image degradations affect encoder-specific recognizability. TransFIRA+ extends this framework to cross-camera multimodal person identification in untrimmed surveillance video. In this setting, body recognizability scores saturate within a narrow range, making direct weighting effectively uniform. We introduce a post-hoc sigmoid calibration that transforms these saturated scores into informative weights without retraining. After calibration, the BRIAR-trained TransFIRA predictor improves fused face and body identification on two unseen multi-camera datasets, outperforming detector-confidence weighting and uncalibrated CCS-rank weighting. Together, TransFIRA and TransFIRA+ demonstrate that encoder-grounded recognizability generalizes across biometric modalities and recognition settings.
\end{abstract}


\section{Introduction}

Biometric recognition in unconstrained settings such as surveillance, long-form video, and web imagery must contend with severe variation in pose, blur, illumination, occlusion, and distance.
Under such conditions, recognition performance cannot be inferred reliably from superficial measures of image quality.
What ultimately matters is whether a sample is \emph{recognizable} to the deployed encoder, since even a few unrecognizable frames can corrupt a template and compromise downstream matching.
Closing this gap between appearance-based quality and encoder-specific recognizability is the central challenge of \emph{face image quality assessment (FIQA)} and, more broadly, of biometric quality assessment.

Despite progress in face recognition~\cite{wang2018cosface, deng2019arcface, narayan2025petalface, nair2025improved}, FIQA methods remain limited in generalization and interpretability.
Hand-crafted cues such as sharpness, pose, or illumination~\cite{raghavendra2014automatic, lijun2019multi, gao2007standardization, IconicityFace} often fail to reflect recognizability in the encoder's embedding space.
Regression- and uncertainty-based pipelines~\cite{8987255, SDD-FIQA2021, DBLP:conf/cvpr/TerhorstKDKK20} rely on curated labels, auxiliary supervision, or costly inference, making them indirect and resource-heavy.
Model-based methods~\cite{meng2021magface, Boutros_2023_CVPR, DBLP:conf/cvpr/KolfDB22} couple FIQA to specific backbones, restricting adaptability, while generative and multimodal pipelines~\cite{babnikIJCB2023, babnikTBIOM2024, Ou_2024_CVPR} improve performance but add complexity without grounding in the encoder's decision geometry.
Crucially, none provides a geometrically principled, encoder-specific definition of recognizability, leaving their predictions as proxies rather than direct measures.

In the conference version of this work~\cite{Tu2026TransFIRA}, we introduced \textbf{Transfer Learning for Face Image Recognizability Assessment (TransFIRA)}, the first framework to adapt pretrained recognition encoders for recognizability prediction without human quality annotations or external IQA supervision.
TransFIRA derives supervision directly from encoder embeddings through \emph{class-center similarity} (CCS) and \emph{class-center angular separation} (CCAS), producing encoder-specific scores tied to the model's discrimination geometry.
These predictions enable \textbf{recognizability-informed aggregation}: CCS-based weighting emphasizes reliable embeddings, while the natural cutoff $CCAS>0$ filters frames predicted to be unrecognizable.
In addition to surpassing prior FIQA methods~\cite{meng2021magface, Boutros_2023_CVPR, babnikIJCB2023} on challenging benchmarks such as BRIAR Protocol~3.1~\cite{cornett2023expanding} and IJB-C~\cite{maze2018iarpa}, TransFIRA extends to body recognition through sigmoid calibration and supports encoder-grounded analysis of how degradations such as blur affect recognizability.

Biometric recognition systems deployed in unconstrained settings must operate beyond verification of curated template pairs.
In long-range surveillance, a subject may appear across multiple camera tracks containing hundreds of frames each.
Faces are often small or absent, so identification relies primarily on body cues and uses face cues when available~\cite{liu2024farsight,corona2021meva,davila2023mevid}.
Robust deployment requires recognizability prediction that transfers across domains, effective template aggregation, and filtering that preserves match coverage.

In \textbf{TransFIRA+}, we extend our conference paper~\cite{Tu2026TransFIRA} by adapting TransFIRA to multimodal cross-camera person identification in untrimmed surveillance video.
Specifically, we analyze weight saturation, introduce post-hoc sigmoid calibration, formulate face and body fusion, and evaluate cross-domain transfer, calibrated aggregation, and the filtering--coverage trade-off on two unseen multi-camera datasets.
Fig.~\ref{fig:teaser} illustrates our framework. The first contribution summarizes TransFIRA~\cite{Tu2026TransFIRA}; the remaining two are new to this article:
\begin{itemize}
\item \textbf{Encoder-grounded recognizability assessment.} We derive recognizability labels from embedding geometry without quality annotations and use the predicted scores for weighted aggregation and principled filtering, achieving state-of-the-art face verification on BRIAR and IJB-C and enabling encoder-grounded degradation analysis.
\item \textbf{Calibrated recognizability weights.} We show that saturated body scores make direct weighting effectively uniform and introduce post-hoc sigmoid calibration that produces informative weights without retraining.
\item \textbf{Multimodal video identification.} We demonstrate cross-domain transfer of a BRIAR-trained body predictor on unseen multi-camera datasets, where calibrated weighting improves fused face and body identification.
\end{itemize}
\section{Related Work}

Research in face image quality assessment (FIQA) has evolved along four main directions.
Analytical methods estimate quality from hand-crafted image statistics such as sharpness, pose, or illumination~\cite{raghavendra2014automatic, lijun2019multi, gao2007standardization, IconicityFace}.
These approaches generalize poorly because visual degradations do not consistently align with recognizability, leaving their scores detached from the decision geometry of modern recognition models.

Regression- and uncertainty-based methods learn mappings from images or embeddings to continuous quality scores.
FaceQNet~\cite{8987255} learns from automatically generated recognition-based labels, SDD-FIQA~\cite{SDD-FIQA2021} models similarity-distribution distance, and SER-FIQ~\cite{DBLP:conf/cvpr/TerhorstKDKK20} estimates reliability through Monte Carlo dropout.
Although correlated with accuracy, these methods depend on curated labels, auxiliary supervision, or stochastic sampling, making their quality estimates indirect and, in some cases, computationally expensive.

Model-based methods integrate FIQA directly into recognition training.
MagFace~\cite{meng2021magface} encodes quality through embedding magnitudes, CR-FIQA~\cite{Boutros_2023_CVPR} learns relative classifiability in angular-margin space, and GraFIQs~\cite{DBLP:conf/cvpr/KolfDB22} derives quality from gradient magnitudes induced by batch-normalization discrepancies.
Although effective, these methods are coupled to particular backbones and training objectives, limiting transfer to arbitrary pretrained encoders.

Generative and multimodal methods offer heavier alternatives.
CLIB-FIQA~\cite{Ou_2024_CVPR} uses multi-factor calibration in a vision--language framework, while DifFIQA~\cite{babnikIJCB2023} and eDifFIQA~\cite{babnikTBIOM2024} perturb images with diffusion models to estimate recognizability.
These pipelines remain indirect proxies for encoder behavior and incur substantial computational cost.

Beyond single images, set- and video-based methods aggregate frames using learned quality weights or attention~\cite{liu2017qan,yang2017nan}, while video person re-identification commonly pools features within tracks~\cite{ye2022reidsurvey}.
Because these strategies are trained jointly with or specifically for a recognition model, they are not readily portable to other pretrained encoders.
At the multimodal level, whole-body systems such as FarSight~\cite{liu2024farsight} combine face and body cues at long range through score-level fusion~\cite{ross2003information}.

In contrast, TransFIRA~\cite{Tu2026TransFIRA} derives recognizability labels directly from class-center geometry and adapts pretrained encoders through transfer learning with a lightweight prediction head, requiring neither quality annotations nor backbone-specific redesign.
Its encoder-specific scores support decision-boundary-aligned filtering, template weighting, and degradation analysis.
In TransFIRA+, we address score saturation and evaluate calibrated recognizability weighting for multimodal cross-camera video identification under domain shift.
\section{Method}

\textbf{TransFIRA} adapts a pretrained recognition encoder to predict \emph{recognizability}, defined in terms of an image embedding's alignment with the correct identity center and separation from competing identities.
These geometric relationships yield encoder-specific training labels without human quality annotations, external IQA supervision, or visual proxies such as blur or illumination.
We finetune a pretrained backbone with a lightweight regression head to predict these labels from raw images, producing encoder-specific scores that guide frame selection and weighting during template construction.
For body encoders with saturated similarities, sigmoid calibration restores useful weight variation before score-level fusion combines face and body comparison scores.

\subsection{Image Recognizability}
\label{sec:method:recognizability}

To supervise recognizability prediction, we derive labels directly from encoder embeddings by comparing each image embedding with its own and competing identity centers.
This grounds recognizability in the chosen encoder's embedding space, ensuring that the labels reflect its discrimination geometry rather than superficial proxies such as blur or illumination.
The resulting supervision therefore requires only the recognition encoder and identity labels, not human quality judgments, handcrafted measures, or external IQA supervision.

Let $\mathcal{D} \,{=}\, \{(x_i,\,y_i)\}_{i=1}^{N}$ contain $N$ images $x_i \in \mathbb{R}^{H \times W \times 3}$ with identity labels $y_i \in \{1,\dots,K\}$, where $K$ is the number of identities.
A fixed recognition encoder $\phi$ maps each image to a $d$-dimensional embedding:
\begin{equation}
z_i = \phi(x_i).
\end{equation}
We measure the recognizability of $x_i$ relative to the \textbf{class center} of its identity.
With a separate gallery set $\mathcal{G}$, class centers use only gallery embeddings:
\begin{equation}
\mu_j \;=\; \frac{1}{n_j^{\mathcal{G}}} \sum_{\substack{i \in \mathcal{G} \\ y_i = j}} z_i,
\end{equation}
where $n_j^{\mathcal{G}}$ is the number of gallery samples for class $j$.
If no gallery--probe split exists, the center is computed from all $n_j$ embeddings of identity $j$ in $\mathcal{D}$:
\begin{equation}
\mu_j \;=\; \frac{1}{n_j} \sum_{i:\, y_i = j} z_i.
\end{equation}

We define the \textbf{Class Center Angular Similarity (CCS)} as the cosine similarity between $z_i$ and its identity center:
\begin{equation}
CCS_{x_i} = \frac{z_i^\top \mu_{y_i}}{\|z_i\|_2 \, \|\mu_{y_i}\|_2},
\end{equation}
which measures alignment with the correct identity.
The \textbf{Nearest Nonmatch Class Center Angular Similarity (NNCCS)} is the largest similarity to an impostor center:
\begin{equation}
NNCCS_{x_i} = \max_{\,j \neq y_i} \frac{z_i^\top \mu_j}{\|z_i\|_2 \, \|\mu_j\|_2},
\end{equation}
which measures resemblance to the most confusable competing identity.
Fig.~\ref{fig:teaser} illustrates how CCS and NNCCS are computed and distributed on BRIAR Protocol~3.1~\cite{cornett2023expanding}.
Finally, their difference defines the \textbf{Class Center Angular Separation (CCAS)}, the similarity margin between the correct and nearest impostor centers:
\begin{equation}
CCAS_{x_i} = CCS_{x_i} - NNCCS_{x_i}.
\end{equation}
Large positive values indicate clear separation from impostor centers, negative values indicate nearest-center misclassification, and values near zero indicate ambiguity.
Under the nearest-class-center decision rule, $CCAS_{x_i}>0$ means that the correct identity center is more similar than every impostor center, making zero a natural, parameter-free threshold for recognizability-based filtering.

Although CR-FIQA's \textbf{Certainty Ratio (CR)}~\cite{Boutros_2023_CVPR} is also a function of CCS and NNCCS, its ratio form does not preserve the angular margin.
The added constant in its denominator changes the scale nonuniformly.
For example, two samples with $(CCS,NNCCS)=(0.9,0.8)$ and $(0.3,0.2)$ are equally well separated ($CCAS=0.1$), yet their CR values differ sharply ($0.5$ versus $0.25$).
By contrast, CCS measures alignment with the correct center and CCAS measures separation from impostors, so $CCAS>0$ corresponds exactly to correct recognition under the nearest-center rule.
The definition depends only on embeddings and class centers and therefore applies across modalities, although saturated score distributions may require calibration before weighting.

\subsection{Recognizability Prediction Network}
\label{sec:method:network}

Because the recognizability labels are deterministic functions of the fixed encoder embeddings, we precompute them for training.
We then initialize a trainable prediction encoder $\phi_\theta$ from $\phi$ and attach a lightweight linear \textbf{recognizability prediction head} $h_\psi$.
For an input $x_i$, the prediction branch outputs a two-dimensional recognizability vector:
\begin{equation}
\hat{\mathbf{r}}_i = \big[\hat{CCS}_{x_i},\, \hat{CCAS}_{x_i}\big]^\top = h_\psi(\phi_\theta(x_i)).
\end{equation}
We optimize $\phi_\theta$ and $h_\psi$ end-to-end using mean squared error against labels computed from the fixed recognition encoder $\phi$:
\begin{align}
\mathbf{r}_i &\equiv [CCS_{x_i},\, CCAS_{x_i}]^\top, \\
\mathcal{L} &= \frac{1}{N}\sum_{i=1}^N \|\hat{\mathbf{r}}_i - \mathbf{r}_i \|_2^2 .
\end{align}

By jointly finetuning $\phi_\theta$ and $h_\psi$ to regress CCS and CCAS, TransFIRA integrates recognizability prediction with the chosen encoder rather than treating it as a detached add-on.
Training the recognizability predictor leaves the recognition representation unchanged, while the pretrained recognition encoder $\phi$ remains fixed and continues to produce embeddings for matching.
The same image may be highly recognizable to one encoder yet ambiguous to another because their embedding geometries differ.
Accordingly, predicted scores reflect the deployed encoder's behavior rather than generic image quality.
Appendix~\ref{sec:experiments:ablation} presents ablations of training scope and supervision targets, showing that end-to-end finetuning with pretrained initialization performs best and that CCS provides the most stable supervision.
Section~\ref{sec:experiments:explainability} further demonstrates that per-image predictions support encoder-grounded analysis of reliability under perturbations.

\subsection{Recognizability-Informed Template Aggregation}
\label{sec:method:aggregation}

Image-level recognizability predictions are especially valuable for template aggregation.
In real-world surveillance and template-based benchmarks such as BRIAR~\cite{cornett2023expanding} and IJB-C~\cite{maze2018iarpa}, many heterogeneous frames are combined into a single subject representation.
Even a few unrecognizable images can corrupt the aggregate and degrade matching accuracy.
Uniform averaging gives every frame equal influence, leaving the representation vulnerable to such inputs.
TransFIRA instead filters and weights frames using predicted recognizability, producing cleaner, more discriminative templates.

We first apply \textbf{recognizability-based filtering}, retaining only images with $\hat{CCAS}_{x_i}>0$.
Under the nearest-center rule, each retained image is predicted to favor its correct identity over every impostor.
The threshold is therefore parameter-free and intrinsic to the deployed encoder.
We then perform \textbf{recognizability-weighted aggregation}, weighting each retained embedding by its predicted CCS:
\begin{equation}
\hat{\mu}^{(s)} =
\frac{\sum_{k \in \mathcal{T}^{(s)}} w_k\, z_k}
     {\sum_{k \in \mathcal{T}^{(s)}} w_k}, \qquad w_k = \hat{CCS}_{x_k},
\label{eq:weighted_template}
\end{equation}
where $\hat{\mu}^{(s)}$ is the template for subject $s$ and $\mathcal{T}^{(s)}$ is the set of retained images.
The evaluated weighting rules satisfy $\sum_{k\in\mathcal{T}^{(s)}}w_k>0$ whenever $\mathcal{T}^{(s)}$ is nonempty; if no image survives filtering, no template is produced.
Because CCS measures proximity to the class center, weighting emphasizes samples that are compactly embedded within their class.

Filtering removes frames predicted to be misrecognized, while weighting increases the influence of the most reliable retained frames.
The two operations are geometrically complementary: CCAS filtering targets inter-class separation, whereas CCS weighting favors intra-class compactness.
Section~\ref{sec:experiments:aggregation} evaluates their individual and combined effects and compares the natural cutoff with CR-FIQA filtering.

\subsection{Calibrated Recognizability Weighting}
\label{sec:method:calibration}

Weighted aggregation departs from uniform averaging only when normalized frame weights vary.
For a template with retained-frame set $\mathcal{T}$, we define normalized weights by $\bar{w}_k{=}w_k/\sum_{\ell\in\mathcal{T}}w_\ell$ and quantify their concentration using the effective-observation fraction
\begin{equation}
\rho(\mathcal{T}) = \frac{1}{|\mathcal{T}| \sum_{k \in \mathcal{T}} \bar{w}_k^2} \in \left(0, 1\right],
\label{eq:effective_fraction}
\end{equation}
This quantity is Kish's effective sample size~\cite{kish1965survey} divided by $|\mathcal{T}|$.
Substantial variation in CCS produces nonuniform weights that emphasize more recognizable frames.
The effective-observation fraction captures this concentration: uniform weights give $\rho=1$, whereas smaller values indicate that fewer frames dominate the template.
If CCS instead concentrates within a narrow range, the normalized weights approach $\rho=1$ and weighting becomes effectively uniform.

\begin{figure}[t]
  \centering
  \includegraphics[width=\linewidth]{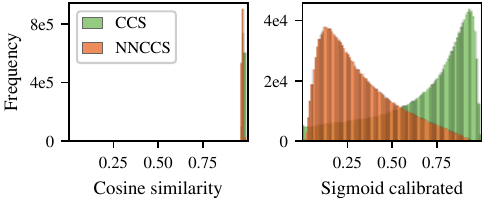}
  \caption{\textbf{Sigmoid calibration of BRIAR~\cite{cornett2023expanding} body recognizability with a SemReID~\cite{huang2023self} encoder trained on the same dataset.}
  Left: raw CCS and NNCCS concentrate near one (mean $0.97$), making direct weighting effectively uniform.
  Right: calibration maps CCS toward one and NNCCS toward zero, spreading the similarities across $[0,1]$ (combined mean $0.50$).}
  \label{fig:sigmoid_calibration}
\end{figure}

We resolve this by repurposing \textbf{sigmoid calibration}~\cite{platt1999probabilistic,guo2017calibration} to shape aggregation weights rather than estimate probabilities.
Using only training identities, we assign binary label one to each CCS value and zero to each NNCCS value, then fit the monotone logistic map
\begin{equation}
g(s) = \operatorname{sigmoid}\!\left(\alpha \, \frac{s - \mu_s}{\sigma_s} + \beta\right), \qquad \alpha \geq 0,
\label{eq:calibration}
\end{equation}
where $\mu_s$ and $\sigma_s$ denote the mean and population standard deviation of the combined CCS and NNCCS values.
We estimate $(\alpha,\beta)$ by minimizing binary cross-entropy with a quadratic slope penalty of $10^{-6}$.
Fig.~\ref{fig:sigmoid_calibration} illustrates how the fitted map spreads saturated similarities over $[0,1]$ by mapping CCS toward one and NNCCS toward zero.

Sigmoid calibration can be applied during target construction or after prediction.
In \emph{target calibration}, $g$ transforms CCS and NNCCS before the training targets are formed, yielding $\tilde{CCS}=g(CCS)$ and $\tilde{CCAS}=g(CCS)-g(NNCCS)$.
In \emph{post-hoc calibration}, the predictor is trained on raw labels and $g$ is applied at inference without retraining:
\begin{align}
\tilde{CCS}_{x} &= g\big(\hat{CCS}_{x}\big), \\
\tilde{CCAS}_{x} &= g\big(\hat{CCS}_{x}\big) - g\big(\hat{CCS}_{x} - \hat{CCAS}_{x}\big),
\end{align}
where $\hat{CCS}_{x}-\hat{CCAS}_{x}$ is the implied nearest-nonmatch similarity.
Calibrated CCAS is therefore the difference between calibrated match and nonmatch similarities, not a sigmoid of the raw margin.
For calibrated aggregation, we set $w_k=\tilde{CCS}_{x_k}$ in Eq.~\eqref{eq:weighted_template}.
The fitted solution has $\alpha>0$, making $g$ strictly increasing and preserving CCS ordering and CCAS sign.
Thus, the $CCAS>0$ threshold remains unchanged, and gains over CCS-rank weighting arise solely from calibrating relative weight magnitudes.

\begin{table*}[t]
\centering
\caption{\textbf{Face template verification results.} Baseline methods use FIQA scores for weighted aggregation;\\\textbf{CCAS Filter} and \textbf{Filter + Weight} apply the $CCAS>0$ cutoff, while \emph{Average} denotes uniform aggregation.\\\rankkey{}.}
\renewcommand{\arraystretch}{1.2}
\resizebox{\textwidth}{!}{%
    \begin{tabular}{l|ccc|ccc|ccc|ccc}
    \toprule
    \multirow{3}{*}{\textbf{Method}} & \multicolumn{6}{c|}{\textbf{BRIAR Protocol 3.1~\cite{cornett2023expanding}: TAR at Fixed FMR}} & \multicolumn{6}{c}{\textbf{IJB-C~\cite{maze2018iarpa}: TAR at Fixed FMR}} \\
     & \multicolumn{3}{c|}{\textbf{CosFace (BRIAR)~\cite{wang2018cosface, cornett2023expanding}}} & \multicolumn{3}{c|}{\textbf{ArcFace (WebFace)~\cite{guo2021insightface, zhu2021webface260m}}} & \multicolumn{3}{c|}{\textbf{CosFace (BRIAR)~\cite{wang2018cosface, cornett2023expanding}}} & \multicolumn{3}{c}{\textbf{ArcFace (WebFace)~\cite{guo2021insightface, zhu2021webface260m}}} \\
     & $10^{-3}$ & $10^{-4}$ & $10^{-6}$ & $10^{-3}$ & $10^{-4}$ & $10^{-6}$ & $10^{-3}$ & $10^{-4}$ & $10^{-6}$ & $10^{-3}$ & $10^{-4}$ & $10^{-6}$ \\
    \midrule
    Average (Baseline) & 0.9077 & 0.8423 & 0.7500 & 0.4269 & 0.1923 & 0.0846 & 0.9675 & 0.9168 & 0.2440 & 0.9734 & 0.9564 & 0.7881 \\
    FaceQNet~\cite{8987255} & 0.9077 & 0.8500 & 0.7577 & 0.4385 & 0.2385 & 0.0923 & 0.9642 & 0.8912 & 0.2224 & 0.9717 & 0.9529 & 0.7567 \\
    MagFace~\cite{meng2021magface} & 0.9077 & 0.8423 & 0.7538 & 0.4385 & 0.2154 & 0.0923 & 0.9694 & 0.9280 & 0.2504 & 0.9736 & 0.9575 & 0.7970 \\
    SDD-FIQA~\cite{SDD-FIQA2021} & 0.9038 & 0.8462 & 0.7538 & 0.4269 & 0.2154 & 0.0923 & 0.9708 & 0.9361 & 0.2525 & 0.9738 & 0.9590 & 0.7977 \\
    CR-FIQA (S)~\cite{Boutros_2023_CVPR} & 0.9038 & 0.8462 & 0.7577 & 0.4538 & 0.2577 & 0.1077 & 0.9721 & 0.9434 & 0.2522 & 0.9742 & 0.9595 & 0.7966 \\
    CR-FIQA (L)~\cite{Boutros_2023_CVPR} & 0.9231 & 0.8731 & 0.8077 & 0.5346 & 0.3231 & 0.1654 & \cellthird{0.9800} & \cellthird{0.9617} & \cellthird{0.3711} & \cellsecond{0.9811} & \cellbest{0.9700} & \cellbest{0.8726} \\
    DifFIQA (R)~\cite{babnikIJCB2023} & 0.9423 & 0.8769 & 0.8269 & 0.6192 & \cellthird{0.4885} & \cellthird{0.3885} & 0.9720 & 0.9447 & 0.2499 & 0.9739 & 0.9596 & 0.7972 \\
    eDifFIQA (S)~\cite{babnikTBIOM2024} & 0.8885 & 0.8462 & 0.7962 & 0.5500 & 0.3731 & 0.2808 & 0.9720 & 0.9454 & 0.2531 & 0.9739 & 0.9597 & 0.7958 \\
    eDifFIQA (M)~\cite{babnikTBIOM2024} & 0.6769 & 0.5923 & 0.4423 & 0.3962 & 0.3000 & 0.2115 & 0.9717 & 0.9482 & 0.2518 & 0.9740 & 0.9597 & 0.7956 \\
    eDifFIQA (L)~\cite{babnikTBIOM2024} & 0.8500 & 0.8000 & 0.7154 & 0.6154 & 0.4692 & 0.3577 & 0.9721 & 0.9468 & 0.2529 & 0.9739 & 0.9599 & 0.7971 \\
    GraFIQs (S)~\cite{DBLP:conf/cvpr/KolfDB22} & 0.9077 & 0.8423 & 0.7500 & 0.4269 & 0.1923 & 0.0846 & 0.9676 & 0.9175 & 0.2441 & 0.9735 & 0.9565 & 0.7884 \\
    GraFIQs (L)~\cite{DBLP:conf/cvpr/KolfDB22} & 0.9077 & 0.8423 & 0.7500 & 0.4269 & 0.1923 & 0.0846 & 0.9675 & 0.9169 & 0.2440 & 0.9734 & 0.9564 & 0.7881 \\
    CLIB-FIQA~\cite{Ou_2024_CVPR} & 0.9154 & 0.8654 & 0.7885 & 0.5038 & 0.3115 & 0.1769 & 0.9714 & 0.9428 & 0.2513 & 0.9740 & 0.9594 & 0.7979 \\
    \midrule
    \textbf{Ours (CCAS Filter)} & \cellsecond{0.9846} & \cellsecond{0.9231} & \cellsecond{0.8846} & \cellsecond{0.8715} & \cellsecond{0.7510} & \cellsecond{0.5703} & \cellsecond{0.9811} & \cellsecond{0.9697} & \cellsecond{0.8334} & \cellbest{0.9811} & \cellthird{0.9685} & \cellthird{0.8712} \\
    \textbf{Ours (CCS Weight)} & \cellthird{0.9615} & \cellthird{0.9038} & \cellthird{0.8462} & \cellthird{0.6269} & 0.4308 & 0.2692 & 0.9718 & 0.9447 & 0.2512 & 0.9737 & 0.9578 & 0.7970 \\
    \textbf{Ours (Filter + Weight)} & \cellbest{0.9885} & \cellbest{0.9269} & \cellbest{0.8885} & \cellbest{0.8715} & \cellbest{0.7510} & \cellbest{0.5984} & \cellbest{0.9815} & \cellbest{0.9711} & \cellbest{0.8601} & \cellthird{0.9809} & \cellsecond{0.9693} & \cellsecond{0.8724} \\
    \bottomrule
    \end{tabular}%
}
\label{tab:tar_results}
\end{table*}

\section{Face and Body Verification}
\label{sec:experiments}

We evaluate TransFIRA through face and body template verification and image-level face recognizability.
Our experiments compare recognizability-informed filtering and weighting with prior FIQA methods, measure how faithfully predicted scores track encoder geometry, and test whether the framework extends from face to body recognition.

\subsection{Evaluation Benchmarks}
\label{sec:experiments:datasets}

We use two complementary template-based benchmarks.
\textbf{BRIAR Protocol~3.1}~\cite{cornett2023expanding} contains 615 subjects~\cite{narayan2025petalface} and emphasizes long-range video surveillance under severe degradation.
\textbf{IJB-C}~\cite{maze2018iarpa} contains 3,531 subjects drawn from diverse web and video imagery and provides a standard, reproducible benchmark for unconstrained face recognition.
Unlike saturated benchmarks such as LFW~\cite{LFWTech}, BRIAR and IJB-C expose enough variation in recognition difficulty to distinguish FIQA methods.
The private BRIAR distribution provides a realistic test of deployment shift for open-source models, whereas IJB-C supports evaluation with public encoders.
Despite their typically degraded appearance, many BRIAR frames remain recognizable to modern encoders, exposing the gap between generic image quality and encoder-specific recognizability.
Dataset demographics and collection procedures, including IRB oversight and participant consent, are documented in~\cite{yoon2026briar}; all BRIAR subjects whose images appear in this paper also consented to publication.
IJB-C consists of public web imagery collected without individual consent.

\begin{table*}[t]
\centering
\caption{\textbf{Image-level face recognizability results.}\\Spearman correlation (SC) is computed against CCS for every method except \textbf{Ours (CCAS)}, which uses CCAS.\\Higher SC and lower AUC are better. \rankkey{}.
}
\renewcommand{\arraystretch}{1.2}
\resizebox{\textwidth}{!}{%
    \begin{tabular}{l|cccc|cccc|cccc|cccc}
    \toprule
    \multirow{3}{*}{\textbf{Method}} & \multicolumn{8}{c|}{\textbf{BRIAR Protocol 3.1~\cite{cornett2023expanding}: FNMR-ERC AUC}} & \multicolumn{8}{c}{\textbf{IJB-C~\cite{maze2018iarpa}: FNMR-ERC AUC}} \\
     & \multicolumn{4}{c|}{\textbf{CosFace (BRIAR)~\cite{wang2018cosface, cornett2023expanding}}} & \multicolumn{4}{c|}{\textbf{ArcFace (WebFace)~\cite{guo2021insightface, zhu2021webface260m}}} & \multicolumn{4}{c|}{\textbf{CosFace (BRIAR)~\cite{wang2018cosface, cornett2023expanding}}} & \multicolumn{4}{c}{\textbf{ArcFace (WebFace)~\cite{guo2021insightface, zhu2021webface260m}}} \\
     & SC & $10^{-3}$ & $10^{-4}$ & $10^{-6}$ & SC & $10^{-3}$ & $10^{-4}$ & $10^{-6}$ & SC & $10^{-3}$ & $10^{-4}$ & $10^{-6}$ & SC & $10^{-3}$ & $10^{-4}$ & $10^{-6}$ \\
    \midrule
FaceQNet~\cite{8987255} & 0.1589 & 0.3957 & 0.5225 & 0.6903 & 0.1194 & 0.5037 & 0.7606 & 0.8924 & 0.3607 & 0.0143 & 0.0438 & 0.6716 & 0.3677 & 0.0154 & 0.0288 & 0.1239 \\
    MagFace~\cite{meng2021magface} & 0.1367 & 0.4314 & 0.5616 & 0.7336 & 0.0467 & 0.5426 & 0.8054 & 0.9200 & \cellthird{0.5684} & 0.0144 & 0.0439 & \cellthird{0.6663} & \cellthird{0.6030} & 0.0156 & 0.0290 & 0.1237 \\
    SDD-FIQA~\cite{SDD-FIQA2021} & 0.0293 & 0.4513 & 0.5771 & 0.7335 & 0.0112 & 0.5527 & 0.8069 & 0.9148 & 0.5374 & 0.0144 & 0.0441 & 0.6668 & 0.5682 & 0.0156 & 0.0291 & 0.1242 \\
    CR-FIQA (S)~\cite{Boutros_2023_CVPR} & 0.1107 & 0.4199 & 0.5463 & 0.7093 & 0.0570 & 0.5409 & 0.7874 & 0.8975 & 0.4218 & \cellthird{0.0142} & 0.0436 & 0.6700 & 0.4589 & 0.0154 & 0.0287 & 0.1232 \\
    CR-FIQA (L)~\cite{Boutros_2023_CVPR} & 0.4143 & 0.2708 & 0.3839 & 0.5640 & 0.3654 & 0.4387 & 0.6815 & 0.8386 & 0.5571 & 0.0144 & 0.0440 & 0.6670 & 0.5902 & 0.0156 & 0.0290 & 0.1237 \\
    DifFIQA (R)~\cite{babnikIJCB2023} & 0.4416 & 0.2782 & 0.3820 & 0.5490 & \cellthird{0.4657} & \cellthird{0.4375} & \cellthird{0.6622} & \cellthird{0.8194} & 0.4729 & 0.0144 & 0.0440 & 0.6688 & 0.5193 & 0.0156 & 0.0290 & 0.1240 \\
    eDifFIQA (S)~\cite{babnikTBIOM2024} & 0.3610 & 0.2978 & 0.4064 & 0.5763 & 0.2577 & 0.4644 & 0.7035 & 0.8433 & 0.5236 & 0.0144 & 0.0439 & 0.6672 & 0.5663 & 0.0156 & 0.0290 & 0.1236 \\
    eDifFIQA (M)~\cite{babnikTBIOM2024} & 0.2870 & 0.3249 & 0.4267 & 0.5836 & 0.2265 & 0.4517 & 0.6895 & 0.8316 & 0.5277 & 0.0146 & 0.0444 & 0.6683 & 0.5715 & 0.0158 & 0.0292 & 0.1248 \\
    eDifFIQA (L)~\cite{babnikTBIOM2024} & 0.4377 & 0.2840 & 0.3855 & 0.5489 & 0.4415 & 0.4407 & 0.6658 & 0.8198 & 0.5344 & \cellsecond{0.0141} & \cellbest{0.0433} & 0.6674 & 0.5803 & \cellthird{0.0153} & \cellthird{0.0286} & \cellsecond{0.1229} \\
    GraFIQs (S)~\cite{DBLP:conf/cvpr/KolfDB22} & 0.0350 & 0.4363 & 0.5646 & 0.7344 & -0.0194 & 0.5325 & 0.8102 & 0.9287 & 0.1050 & 0.0143 & 0.0438 & 0.6793 & 0.0911 & 0.0154 & 0.0288 & 0.1237 \\
    GraFIQs (L)~\cite{DBLP:conf/cvpr/KolfDB22} & 0.1831 & 0.3881 & 0.5059 & 0.6732 & 0.1434 & 0.5074 & 0.7643 & 0.8914 & 0.3307 & \cellthird{0.0142} & 0.0436 & 0.6733 & 0.3440 & 0.0154 & 0.0287 & 0.1232 \\
    CLIB-FIQA~\cite{Ou_2024_CVPR} & \cellthird{0.5004} & \cellthird{0.2536} & \cellthird{0.3591} & \cellthird{0.5340} & 0.4404 & 0.4508 & 0.6758 & 0.8281 & 0.5112 & \cellthird{0.0142} & \cellthird{0.0435} & 0.6677 & 0.5580 & 0.0154 & 0.0287 & 0.1231 \\
    \midrule
    \textbf{Ours (CCAS)} & \cellsecond{0.8230} & \cellsecond{0.1558} & \cellsecond{0.2734} & \cellbest{0.4752} & \cellbest{0.6568} & \cellsecond{0.4071} & \cellsecond{0.6335} & \cellbest{0.8068} & \cellbest{0.6245} & 0.0144 & 0.0442 & \cellsecond{0.6659} & \cellbest{0.6306} & \cellbest{0.0153} & \cellsecond{0.0285} & \cellthird{0.1230} \\
    \textbf{Ours (CCS)} & \cellbest{0.8566} & \cellbest{0.1536} & \cellbest{0.2722} & \cellsecond{0.4753} & \cellsecond{0.6401} & \cellbest{0.3934} & \cellbest{0.6315} & \cellsecond{0.8076} & \cellsecond{0.6111} & \cellbest{0.0141} & \cellsecond{0.0433} & \cellbest{0.6659} & \cellsecond{0.6245} & \cellsecond{0.0153} & \cellbest{0.0285} & \cellbest{0.1228} \\
    \bottomrule
    \end{tabular}%
}
\label{tab:erc_auc_results}
\end{table*}

\subsection{Recognition Encoders and Training}
\label{sec:experiments:setup}

We evaluate two recognition encoders.
\textbf{CosFace (BRIAR)} is a Swin-B Transformer trained on the private BRIAR dataset with the CosFace objective, achieving state-of-the-art performance on BRIAR Protocol~3.1~\cite{cornett2023expanding,wang2018cosface,liu2021swin}.
\textbf{ArcFace (WebFace)} is the iResNet50 model with the strongest IJB-C performance in the open-source InsightFace repository~\cite{deng2019arcface,guo2021insightface,zhu2021webface260m,maze2018iarpa}.
For each recognition encoder, we initialize the recognizability predictor from its pretrained weights and finetune it end-to-end with a two-output linear head and MSE loss.
Recognizability training uses disjoint identity splits: a subset of BRIAR for the BRIAR experiments and 10,000 WebFace12M identities for IJB-C.
WebFace12M consists of public web imagery collected without individual consent; its demographic data are reported in~\cite{zhu2021webface260m}.
We train each predictor for 50 epochs with AdamW using a 10\textsuperscript{-6} learning rate, 10\textsuperscript{-4} weight decay, and batch size 64, then select the checkpoint with the highest validation Spearman correlation.
Together, these encoders span a domain-matched private system and a reproducible public baseline with distinct architectures and training distributions.

\subsection{Face Template Verification}
\label{sec:experiments:aggregation}

Table~\ref{tab:tar_results} compares recognizability-informed filtering and weighting with existing FIQA methods.
Appendices~\ref{sec:supp:crfiqa} and~\ref{sec:supp:roc} provide a direct comparison with CR-FIQA filtering and the corresponding ROC curves.
On BRIAR Protocol~3.1, CCS weighting is the strongest unfiltered method at every operating point.
The intrinsic $CCAS>0$ filter yields larger gains, and combining filtering with weighting produces the best overall results.
At 10\textsuperscript{-6} FMR, the combined method raises TAR from 0.750 to 0.889 with CosFace and from 0.085 to 0.598 with ArcFace.

On IJB-C, CCS weighting remains competitive for both encoders, while CCAS filtering and the combined variant perform best with CosFace.
With ArcFace, the best TransFIRA variant is tied for strongest or within 0.001 TAR of CR-FIQA across all operating points.
This small gap may reflect training alignment because both ArcFace and CR-FIQA are optimized on WebFace12M.
More broadly, the results suggest that FIQA methods aligned with an encoder's training distribution can gain slight advantages under in-domain evaluation.

Overall, encoder-grounded filtering and weighting deliver their largest gains on BRIAR while remaining competitive on IJB-C; Appendix~\ref{sec:experiments:crossdataset} reports cross-domain results.
\subsection{Image-Level Face Recognizability}
\label{sec:experiments:erc}

Template verification evaluates downstream utility, whereas image-level analysis tests whether predicted scores themselves track the encoder's recognition geometry.
We measure this fidelity using Spearman correlation (SC) with ground-truth recognizability and FNMR error-versus-reject curves (ERCs) at fixed target FMRs.
ERCs measure how effectively rejecting low-scoring images reduces FNMR.
Table~\ref{tab:erc_auc_results} reports SC and ERC area under the curve (AUC); higher SC and lower AUC indicate better alignment.
Appendix~\ref{sec:supp:erc} provides the corresponding curves.

On BRIAR Protocol~3.1, TransFIRA substantially outperforms every baseline.
With CosFace, predicted CCS attains an SC of 0.86 compared with 0.50 for the strongest competing method, and TransFIRA achieves the lowest AUC at every operating point.
ArcFace follows the same trend, with both CCS and CCAS maintaining substantially stronger alignment than existing FIQA scores.

On IJB-C, TransFIRA ranks first or second across nearly every metric and encoder.
CCS and CCAS attain the highest SC and lowest AUC at most operating points; the sole exception is CosFace at 10\textsuperscript{-4} FMR, where eDifFIQA~\cite{babnikTBIOM2024} leads by less than 0.001 AUC.
These image-level results explain the gains in face template verification: TransFIRA ranks images by recognition utility more faithfully than generic quality scores.

\subsection{Body Template Verification}
\label{sec:experiments:body}

\begin{table}[t]
\centering
\caption{\textbf{Body template verification results with a SemReID body encoder~\cite{huang2023self} trained on BRIAR.} Raw scores use the original labels; calibrated scores use the calibration of Section~\ref{sec:method:calibration}.}
\begin{tabular}{l|*{3}{>{\centering\arraybackslash}p{1.2cm}}}
\toprule
\multirow{2}{*}{\textbf{Method}} & \multicolumn{3}{c}{\textbf{BRIAR~\cite{cornett2023expanding}: TAR at Fixed FMR}} \\
 & $10^{-3}$ & $10^{-4}$ & $10^{-6}$ \\
\midrule
Average (Baseline)  & 0.5934 & 0.3278 & 0.0851 \\
\midrule
Raw CCAS Filter  & 0.5934 &  0.3278 &  0.0851 \\
Raw CCS Weight  & 0.5934 &  0.3257  & 0.0851 \\
\hspace{1em} Filter + Weight  &  0.5934 &  0.3257 &  0.0851 \\

Raw CCAS Weight  & 0.6000  & 0.3340  & 0.0851 \\
\hspace{1em} Filter + Weight  & 0.6000 &  0.3340 &  0.0851 \\
\midrule
Calibrated CCAS Filter  & 0.6037 &  0.3485 &  0.0933 \\
Calibrated CCS Weight & 0.6120 &  0.3402  & 0.0934 \\
\hspace{1em} Filter + Weight  &  0.6203  & 0.3465 &  0.0996\\

Calibrated CCAS Weight  & \textbf{0.6452} &  \textbf{0.3693}  & 0.1037 \\
\hspace{1em} Filter + Weight  &  0.6328 &  0.3568  & \textbf{0.1079} \\
\bottomrule
\end{tabular}
\label{tab:body}
\end{table}

We test whether TransFIRA extends beyond faces by applying it to body template verification on BRIAR Protocol~3.1 with a BRIAR-trained SemReID body encoder~\cite{huang2023self}.
Body recognition is especially challenging because clothing changes, pose variation, and occlusion can overwhelm appearance-based similarity.
As in the face experiments, we derive recognizability labels from similarity to the correct and nearest nonmatch class centers.
Fig.~\ref{fig:sigmoid_calibration} shows that raw CCS and NNCCS values concentrate near one, with mean 0.97 and variance $3.69\times10^{-5}$.
We therefore apply the target calibration from Section~\ref{sec:method:calibration}, mapping CCS toward one and NNCCS toward zero.
The calibrated distributions have mean 0.50, variance 0.09, and Brier score 0.1564, restoring useful variation for aggregation and CCAS estimation.

Table~\ref{tab:body} shows that raw CCS and CCAS weighting produce little improvement over uniform aggregation because score saturation makes their normalized weights nearly uniform.
Calibration changes this pattern: calibrated CCAS weighting achieves the highest TAR at 10\textsuperscript{-3} and 10\textsuperscript{-4} FMR, while adding filtering produces the best result at 10\textsuperscript{-6}.
Relative separability measured by CCAS is therefore more informative for body verification than absolute similarity, which saturates.
Section~\ref{sec:video} revisits these choices in video identification, where filtering also affects template coverage.
\section{Multimodal Video Identification}
\label{sec:video}

We evaluate TransFIRA for multimodal cross-camera person identification in untrimmed security video using two unseen VideoLINCS datasets.
Our experiments examine transfer of a BRIAR-trained body recognizability predictor, calibrated weighting, the tradeoff between filtering and template coverage, and face and body fusion.

\subsection{Evaluation Datasets and Protocol}
\label{sec:video:datasets}
\label{sec:video:protocol}

We use two nonpublic IARPA VideoLINCS datasets with frame-level identity annotations for untrimmed security video.
\textbf{\dsone} contains indoor and courtyard scenes captured by six cameras.
Its test split contains $1{,}241$ tracks from $26$ identities, with every track serving as a query.
Tracks contain a median of $205$ body frames; median body and face heights are $239.8$ and $33.4$ pixels.
\textbf{\mevidvl} draws video and identities from MEVA~\cite{corona2021meva} and MEVID~\cite{davila2023mevid} but uses its own frame-level annotations, reviewed identity set, and retrieval protocol, distinguishing it from the public MEVID tracklet benchmark.
Both datasets were collected under IRB oversight with all actors signing consent forms.
Eleven cameras cover roads, parking areas, entrances, and stairwells.
Its test split contains $2{,}433$ tracks and $2{,}413$ queries from $66$ identities.
Tracks contain a median of $98$ body frames; median body and face heights are $97.5$ and $16.4$ pixels.
Both datasets feature occlusion, viewpoint variation, and border truncation.

We divide identities into disjoint development and test sets, assigning $30\%$ to development within the single- and multi-camera strata.
Development data are used only to fit the fusion model and filter thresholds.
Person detections are assigned one-to-one to annotation boxes at $\mathrm{IoU}{\geq}0.5$, and each face detection inherits the track of its associated body.
Every annotated track with at least five frames is included regardless of detection success, retaining detection and feature failures in the evaluation.
In total, the body predictor scores $1.6$M detections on \dsone{} and $8.4$M on \mevidvl{}.

Every track whose identity appears in another camera is queried against all tracks from other cameras in the same split.
We report mean average precision (mAP) and \mbox{Rank-1} accuracy.
Average precision includes every annotated cross-camera positive, even when no template is available; queries with no scored gallery item receive zero.
We obtain unadjusted $95\%$ confidence intervals from $2{,}000$ identity-bootstrap resamples shared across conditions; intervals for differences are bracketed.
Appendix~\ref{sec:appendix:video} provides full matching, bootstrap, and sampling details.

\subsection{Models and Template Construction}
\label{sec:video:implementation}

Body templates use the ECHO-BID body recognition encoder~\cite{metz2025not}, an EVA02-Large backbone~\cite{fang2024eva02} with $448$-pixel input and $512$-dimensional output.
Following Section~\ref{sec:method:network}, we compute raw CCS and CCAS labels from the frozen encoder on $1.61$M BRIAR body chips spanning $1{,}221$ identities.
We use $976$ identities for training and $245$ for validation, then initialize the recognizability predictor from the ECHO-BID weights and finetune it with two linear heads using the optimization settings of Section~\ref{sec:experiments:setup}.
The selected checkpoint achieves validation Spearman correlations of $0.70$ for CCS and $0.56$ for CCAS.
The labels are strongly saturated, with raw validation CCS in $[0.947, 0.998]$.
We fit the calibration $g$ of Eq.~\eqref{eq:calibration} once on the BRIAR training labels and apply it unchanged to both video datasets.
Recognition continues to use embeddings from the frozen ECHO-BID encoder.

Face templates use a $512$-dimensional Swin face encoder~\cite{liu2021swin} with detector-confidence rank weighting, identical across all body conditions.
Each available face and body template is compared by cosine similarity, and the resulting similarities are combined through standard score-level fusion~\cite{ross2003information}.
For a template pair, let $c_f$ and $c_b$ denote face and body similarities standardized by the mean and standard deviation of development-set nonmatch scores; a missing modality contributes zero after standardization.
The fused score is
\begin{equation}
    s=\gamma_0+\gamma_f c_f+\gamma_b c_b,
    \label{eq:video_fusion}
\end{equation}
where the coefficients are estimated using a single class-balanced logistic regression on uniformly weighted development template pairs and then held fixed across all test conditions.
Pairs with neither modality are unscored.
Holding the fusion model fixed ensures that changes in fused accuracy reflect body template construction.

All conditions share the same detections, embeddings, tracks, and galleries and differ only in how body frames are filtered and weighted.
We compare uniform averaging with detector-confidence and CCS weighting under several score-to-weight mappings.
\emph{Rank} weights use the within-track rank of a score, while \emph{track min--max} rescales scores to $[0,1]$ within each track.
\emph{Training-linear} maps the $1$st and $99$th percentiles of training CCS to zero and one with clipping, and \emph{calibrated} weighting uses $\tilde{CCS}$.
Weights are normalized over the frames that survive any filter, and a track with no surviving frames has no template.
When a mapping is undefined because scores are constant under min--max scaling or all weights are clipped to zero under training-linear scaling, the track falls back to uniform weights over the same frames.

\subsection{Cross-Domain Recognizability Transfer}
\label{sec:video:transfer}

\begin{figure}[t]
  \centering
  \includegraphics[width=\linewidth]{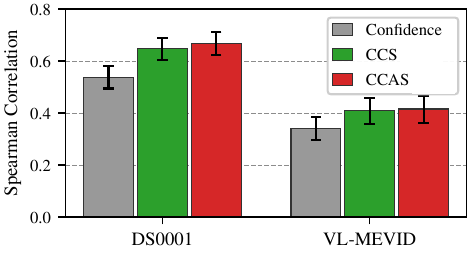}
  \caption{\textbf{Recognizability transfer to unseen video with an ECHO-BID body encoder~\cite{metz2025not} and a TransFIRA predictor trained only on BRIAR~\cite{cornett2023expanding}.} On both unseen datasets, predicted CCS and CCAS track empirical cross-camera angular separation more closely than detector confidence. Bars show mean within-identity Spearman correlation on the \dsone{} and \mevidvl{}~\cite{corona2021meva,davila2023mevid} test splits; error bars show $95\%$ identity-bootstrap intervals.}
  \label{fig:video_transfer}
\end{figure}

To test transfer beyond BRIAR, we compute an empirical cross-camera CCAS label for up to $64$ frames per track.
Each label is the similarity to the correct identity center minus the similarity to the nearest competing center, with all centers formed from body embeddings in other cameras.
The diagnostic covers $59{,}003$ frames from $26$ identities on \dsone{} and $105{,}261$ frames from $66$ identities on \mevidvl{}.
Fig.~\ref{fig:video_transfer} reports Spearman correlations computed within each identity and then averaged across identities.

Predicted CCS and CCAS rank frames better than detector confidence on both datasets: $0.650$ and $0.669$ versus $0.539$ on \dsone{}, and $0.411$ and $0.416$ versus $0.340$ on \mevidvl{}.
Predicted CCAS also identifies frames with positive empirical separation more accurately than confidence, with identity-balanced AUROC of $0.90$ versus $0.83$ on \dsone{} and $0.71$ versus $0.67$ on \mevidvl{}.
On \mevidvl{}, $60.2\%$ of frames retained by $CCAS>0$ have positive empirical separation, compared with $19.1\%$ of rejected frames.
The BRIAR-trained predictor therefore preserves its recognizability signal across unseen cameras and sites without adaptation.

\subsection{Calibrated Template Aggregation}
\label{sec:video:calibration}

Table~\ref{tab:video_results} reports body-only and fused retrieval for each body template construction.
Calibrated CCS achieves the best body-only and fused mAP on both datasets.
Relative to detector-confidence ranks, fused mAP improves by $+2.18$ $[+1.59, +2.90]$ points on \dsone{} and $+1.47$ $[+1.07, +1.84]$ on \mevidvl{}.
On the more difficult \mevidvl{}, calibration also raises fused \mbox{Rank-1} by $+2.90$ $[+1.78, +4.07]$ points.
On \dsone{}, nearly $80\%$ of queries already retrieve a correct match first, leaving less headroom at \mbox{Rank-1}; the change is not statistically resolved.

The comparison with CCS ranks isolates the source of these gains: both methods order frames identically, so calibration differs only in the relative weight magnitudes.
Raw CCS has a median effective-observation fraction $\rho$ above $0.9999$ on both datasets, making direct weighting effectively uniform.
Rank weights impose a fixed linear profile whose effective-observation fraction approaches $\rho{=}0.75$ for long tracks, whereas calibrated weights adapt to each track's recognizability spread, yielding median $\rho$ values of $0.85$ on \dsone{} and $0.80$ on \mevidvl{}.
Calibration therefore improves aggregation by preserving informative score spacing rather than reranking frames.

Other mappings support this interpretation: training-linear scaling nearly matches the sigmoid because both use the training score distribution, but clipping forces $54$ and $104$ test tracks to fall back to uniform weights.
Track min--max scaling discards the absolute scale and is no better than ranks.
Sigmoid calibration thus provides a smooth global mapping without discarding cross-track scale or introducing clipping failures.
Aggregation itself is essential: with CCS rank weighting, $16$ frames per track come within $0.12$ and $0.23$ fused mAP points of full templates on \dsone{} and \mevidvl{}, and CCS ranks outperform confidence ranks at every budget above one frame.

\subsection{Filtering and Template Coverage}
\label{sec:video:filtering}

The lower blocks of Table~\ref{tab:video_results} evaluate the $CCAS>0$ filter.
With detector-confidence ranks, the filter produces only modest changes relative to no filtering, and the intervals include zero.
At matched retention, however, it improves fused mAP over confidence filtering by $+1.10$ $[+0.56, +1.66]$ points on \mevidvl{} and $+0.98$ $[-0.09, +1.98]$ on \dsone{}.
Within the retained frames, calibrated CCS again improves on CCS ranks by $+0.37$ $[+0.17, +0.55]$ points on \dsone{} and $+0.20$ $[+0.10, +0.31]$ on \mevidvl{}.
\begin{table}[t]
\centering
\caption{\textbf{Multimodal video identification with ECHO-BID~\cite{metz2025not}.} Results report body-only and fused mAP, fused \mbox{Rank-1} (R1), and positive coverage (Cov.) as percentages. Only body filtering and weighting vary; face templates and fusion parameters remain fixed. The confidence filter matches the development-set retention of $CCAS>0$.}
\setlength{\tabcolsep}{3pt}
\resizebox{\columnwidth}{!}{
\begin{tabular}{l|cccc|cccc}
\toprule
 & \multicolumn{4}{c|}{\textbf{\dsone}} & \multicolumn{4}{c}{\textbf{\mevidvl{}~\cite{corona2021meva,davila2023mevid}}} \\
\textbf{Body Weighting} & Body & Fused & R1 & Cov. & Body & Fused & R1 & Cov. \\
\midrule
\multicolumn{9}{l}{\emph{No Filter}} \\
Uniform               & 59.29 & 60.00 & 79.21 & 77.2 & 24.19 & 24.27 & 57.56 & 80.1 \\
Confidence Ranks      & 60.22 & 60.79 & 78.89 & 77.2 & 24.82 & 24.90 & 57.85 & 80.1 \\
Confidence Min--Max   & 60.73 & 61.23 & 78.81 & 77.2 & 24.70 & 24.77 & 58.02 & 80.1 \\
CCS Ranks             & 60.69 & 61.23 & 79.13 & 77.2 & 25.40 & 25.49 & 59.51 & 80.1 \\
CCS Track Min--Max    & 60.68 & 61.21 & 79.37 & 77.2 & 25.16 & 25.25 & 59.01 & 80.1 \\
CCS Training-Linear   & 62.42 & 62.86 & 80.18 & 77.2 & 25.86 & 25.99 & 59.88 & 80.1 \\
Calibrated CCS        & \textbf{62.58} & \textbf{62.97} & 79.69 & 77.2 & \textbf{26.28} & \textbf{26.37} & \textbf{60.75} & 80.1 \\
\midrule
\multicolumn{9}{l}{\emph{Matched Confidence Filter}} \\
Confidence Ranks      & 58.72 & 60.11 & 78.73 & 73.1 & 24.22 & 24.41 & 57.52 & 73.4 \\
CCS Ranks             & 59.20 & 60.56 & 79.29 & 73.1 & 24.84 & 25.00 & 58.97 & 73.4 \\
\midrule
\multicolumn{9}{l}{\emph{$CCAS>0$ Filter}} \\
Confidence Ranks      & 59.66 & 61.09 & 80.10 & 70.6 & 25.31 & 25.51 & 58.89 & 72.1 \\
CCS Ranks             & 59.77 & 61.17 & 80.02 & 70.6 & 25.67 & 25.84 & 59.93 & 72.1 \\
Calibrated CCS        & 60.22 & 61.55 & \textbf{80.26} & 70.6 & 25.83 & 26.04 & 60.38 & 72.1 \\
\bottomrule
\end{tabular}
}
\label{tab:video_results}
\end{table}

Although CCAS selects better frames than confidence at matched retention, filtering can still hurt end-to-end identification by removing complete templates.
Relative to ungated calibrated CCS, filtering changes fused mAP by $-1.42$ $[-2.36, -0.48]$ points on \dsone{} and $-0.33$ $[-0.90, +0.12]$ on \mevidvl{}.
A track with no retained frames loses its template while its matches remain in the AP denominator.
Filtering reduces the fraction of scorable positives from $77.2\%$ to $70.6\%$ on \dsone{} and from $80.1\%$ to $72.1\%$ on \mevidvl{}.
Calibrated weighting is therefore preferable for identification, while $CCAS>0$ is better suited when templates retain recognizable frames or abstention is acceptable.

\begin{figure*}[t]
  \centering
  \setlength{\tabcolsep}{0pt}\small
  \begin{tabular}{@{}*{3}{>{\centering\arraybackslash}p{0.23235\linewidth}@{\hspace{0.02354\linewidth}}}>{\centering\arraybackslash}p{0.23235\linewidth}@{}}
  Original & Light Blur & Moderate Blur & Heavy Blur
  \end{tabular}\\[1pt]
  \includegraphics[width=\linewidth,trim=7.20 392.6 7.15 213.0,clip]{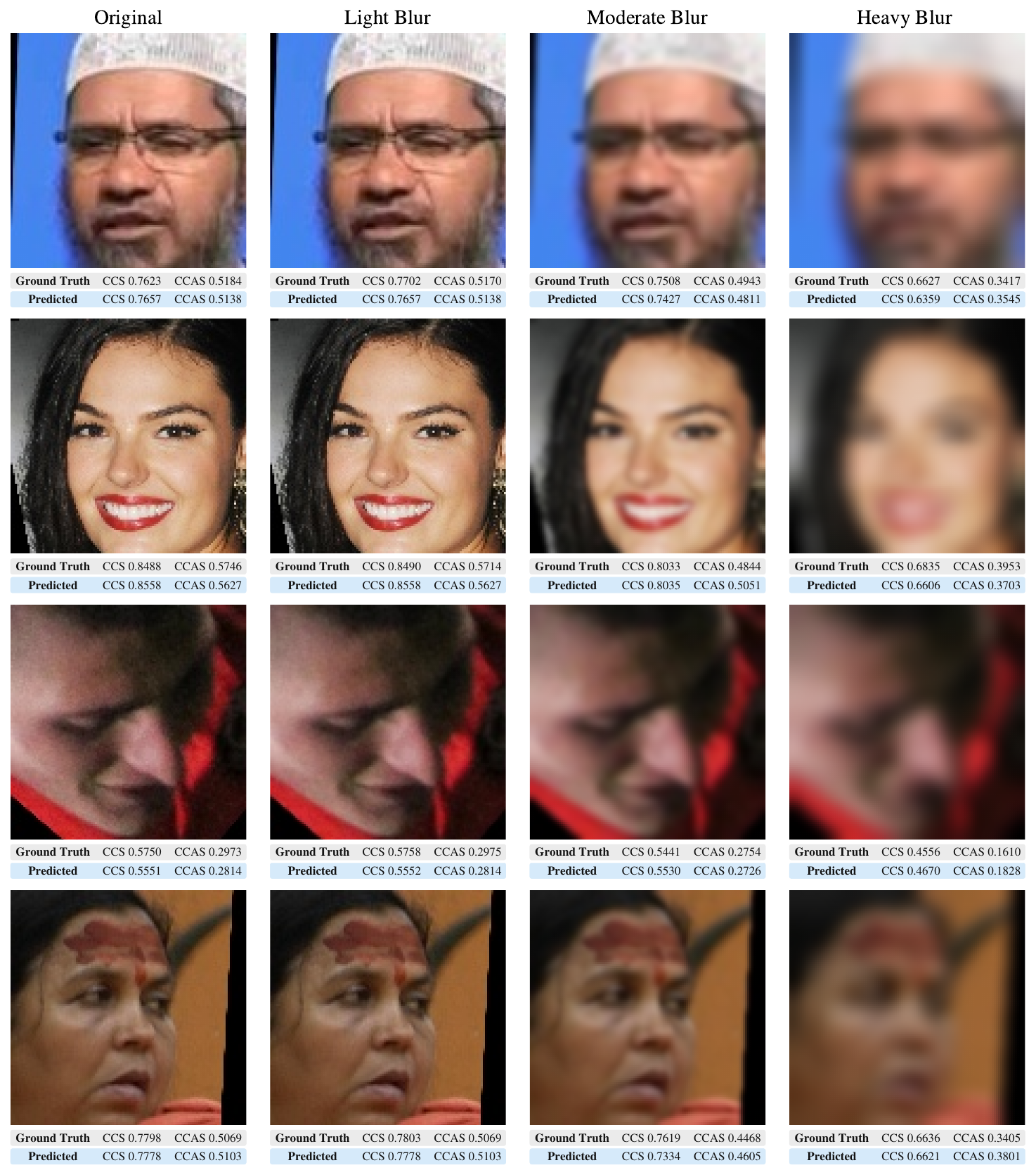}\\
  \caption{\textbf{Recognizability under Gaussian blur with ArcFace~\cite{guo2021insightface,zhu2021webface260m} on IJB-C~\cite{maze2018iarpa}.}
  Light blur leaves recognizability essentially unchanged, while moderate and heavy blur reduce it sharply, and predicted CCS and CCAS track the ground truth at every level.
  Additional examples are shown in Appendix~\ref{sec:appendix:explainability}.}
  \label{fig:explainability}
\end{figure*}

\subsection{Multimodal Fusion}
\label{sec:video:fusion}

Face-only retrieval reaches $41.04\%$ mAP on \dsone{} but only $4.40\%$ on \mevidvl{}, where the median face is $16$ pixels tall and tracks contain a median of $28$ valid face frames.
With calibrated body templates, adding face scores yields modest gains of $+0.39$ $[+0.25, +0.57]$ mAP points on \dsone{} and $+0.09$ $[+0.02, +0.15]$ on \mevidvl{}.
Body cues therefore drive identification at these distances, with face providing limited complementary evidence.
\section{Encoder-Grounded Explainability}
\label{sec:experiments:explainability}

Because TransFIRA requires no ground-truth labels at inference, its predicted scores can provide an encoder-grounded explainability signal for unseen data.
One application is measuring the effect of image degradation on encoder-specific recognizability.
We demonstrate this by applying Gaussian blur to IJB-C images and scoring each image with an ArcFace predictor finetuned on WebFace12M~\cite{guo2021insightface,zhu2021webface260m}.
Table~\ref{tab:explainability} shows that mean predicted CCS and CCAS follow the ground-truth decline as blur increases, with positive Spearman correlations at every level.
Fig.~\ref{fig:explainability} demonstrates that blur affects recognizability nonuniformly, with little change under light blur but sharp reductions under moderate and heavy blur.
Beyond degradation analysis, low-scoring samples can be flagged for recapture or used as hard training examples to improve the robustness of the recognition encoder.
TransFIRA predictions can also be aggregated by camera or capture condition to analyze encoder-specific recognizability.
\section{Limitations and Future Work}
\label{sec:discussion}

TransFIRA retains the full pretrained backbone, so its recognizability predictions remain tied to the deployed encoder's geometry.
Distilling recognizability into a smaller student model could enable lighter deployment but weaken alignment with the encoder.
For video deployment, integrating recognizability prediction with end-to-end tracking would extend TransFIRA throughout the pipeline.
Adding face recognizability would further enable recognizability-aware fusion of face and body cues.
The framework also applies to other biometric modalities such as gait recognition, where frame- and sequence-level scores could select informative viewpoints and gait cycles before aggregation.
Beyond biometrics, encoder-grounded recognizability could support general image recognition by identifying ambiguous samples for data curation, hard-example mining, and selective prediction.
Semantic or prompt-conditioned prototypes could adapt TransFIRA to vision--language models so recognizability reflects the task or prompt.

\begin{table}[t]
    \centering
    \caption{\textbf{Effect of Gaussian blur on recognizability.}\\Mean ground-truth (GT) and predicted (Pred) CCS and CCAS on IJB-C~\cite{maze2018iarpa} with an ArcFace encoder finetuned on WebFace12M~\cite{guo2021insightface,zhu2021webface260m}; SC is their Spearman correlation.}

    \resizebox{\columnwidth}{!}{
    \begin{tabular}{l|ccc|ccc}
    \toprule
    \multirow{2}{*}{\textbf{Blur Level}} & \multicolumn{3}{c|}{\textbf{CCS}} & \multicolumn{3}{c}{\textbf{CCAS}} \\
     & GT & Pred & SC & GT & Pred & SC \\
    \midrule
    None     & 0.6335 & 0.7436 & 0.6171 & 0.3351 & 0.4677 & 0.6594 \\
    Light    & 0.6338 & 0.7434 & 0.6157 & 0.3344 & 0.4676 & 0.6596 \\
    Moderate & 0.6040 & 0.7050 & 0.5554 & 0.2891 & 0.4277 & 0.6314 \\
    Heavy    & 0.4865 & 0.5316 & 0.3143 & 0.0834 & 0.2545 & 0.3962 \\
    \bottomrule
    \end{tabular}}
    \label{tab:explainability}
\end{table}

\section{Conclusion}

We introduced \textbf{TransFIRA}, a lightweight framework that derives encoder-specific recognizability targets from the deployed encoder's embedding geometry without human quality annotations.
Using the predicted recognizability scores for weighted template aggregation and principled filtering achieves state-of-the-art face verification on BRIAR Protocol~3.1 and IJB-C.
Beyond accuracy, the predicted scores provide encoder-grounded explainability for individual samples and deployment conditions.
In \textbf{TransFIRA+}, we extend this approach to multimodal cross-camera identification in untrimmed video.
Post-hoc sigmoid calibration converts saturated body scores into effective aggregation weights without retraining.
The BRIAR-trained predictor transfers to two unseen multi-camera datasets, where these weights improve fused face and body identification.
Together, TransFIRA and TransFIRA+ move beyond visual quality to encoder-specific recognizability, establishing a geometry-driven basis for accurate, interpretable recognition across modalities and settings.

\section*{Acknowledgments}

This research is based upon work supported in part by the Office of the Director of National Intelligence (ODNI), Intelligence Advanced Research Projects Activity (IARPA), via IARPA R\&D Contracts No.~2022-21102100005 and No.~56000026C0026. The views and conclusions contained herein are those of the authors and should not be interpreted as necessarily representing the official policies, either expressed or implied, of ODNI, IARPA, or the U.S. Government. The U.S. Government is authorized to reproduce and distribute reprints for governmental purposes notwithstanding any copyright annotation therein.

\clearpage
\bibliographystyle{IEEEtran}
\bibliography{egbib}

\clearpage

\appendices

\section{Training Ablation}
\label{sec:experiments:ablation}

Table~\ref{tab:component_ablation} ablates the predictor training protocol of Sections~\ref{sec:method:network} and~\ref{sec:experiments:setup}, showing that end-to-end finetuning from pretrained weights outperforms both random initialization and head-only finetuning.
Pretrained models converge after one epoch, whereas random initialization requires roughly 20 epochs, demonstrating the efficiency of transfer learning.
The table also compares CCS, CCAS, and the Certainty Ratio introduced by CR-FIQA~\cite{Boutros_2023_CVPR} as prediction targets computed from the same embeddings.
CCS provides the most stable supervision, while CCAS supplies the margin used for filtering; CR's ratio form gives slightly less consistent supervision.

\section{Additional Verification Results}
\label{sec:appendix:verification}
\subsection{Comparison with CR-FIQA Filtering}
\label{sec:supp:crfiqa}

CR-FIQA~\cite{Boutros_2023_CVPR} defines $CR{=}CCS/(NNCCS+1+\epsilon)$ as its Certainty Ratio, with $\epsilon=10^{-9}$, but does not prescribe a filtering threshold.
A natural candidate is $CR>0.5$, the midpoint of its $[0,1]$ range, which requires the genuine similarity to exceed half of the shifted nonmatch similarity.
The added $1+\epsilon$ shifts the scale arbitrarily, so this threshold has no intrinsic connection to the decision geometry; by contrast, $CCAS>0$ is exactly the condition $CCS>NNCCS$ that defines correct recognition.
Table~\ref{tab:filtering_comparison} extends the filtering analysis of Section~\ref{sec:experiments:aggregation} and shows that this geometric distinction matters.
Filtering with $CR>0.5$ adds little beyond weighting alone, whereas $CCAS>0$ improves TAR on its own and, combined with weighting, consistently outperforms $CR>0.5$ across both backbones on BRIAR and IJB-C.
With CosFace on BRIAR, for example, $CCAS>0$ filtering raises TAR at $10^{-3}$ FMR from $0.9077$ to $0.9846$, and adding CCS weighting raises it further to $0.9885$.

\subsection{ROC and ERC Curves}
\label{sec:supp:roc}
\label{sec:supp:erc}

\begin{figure*}[t!]
  \centering
  \includegraphics[width=\linewidth]{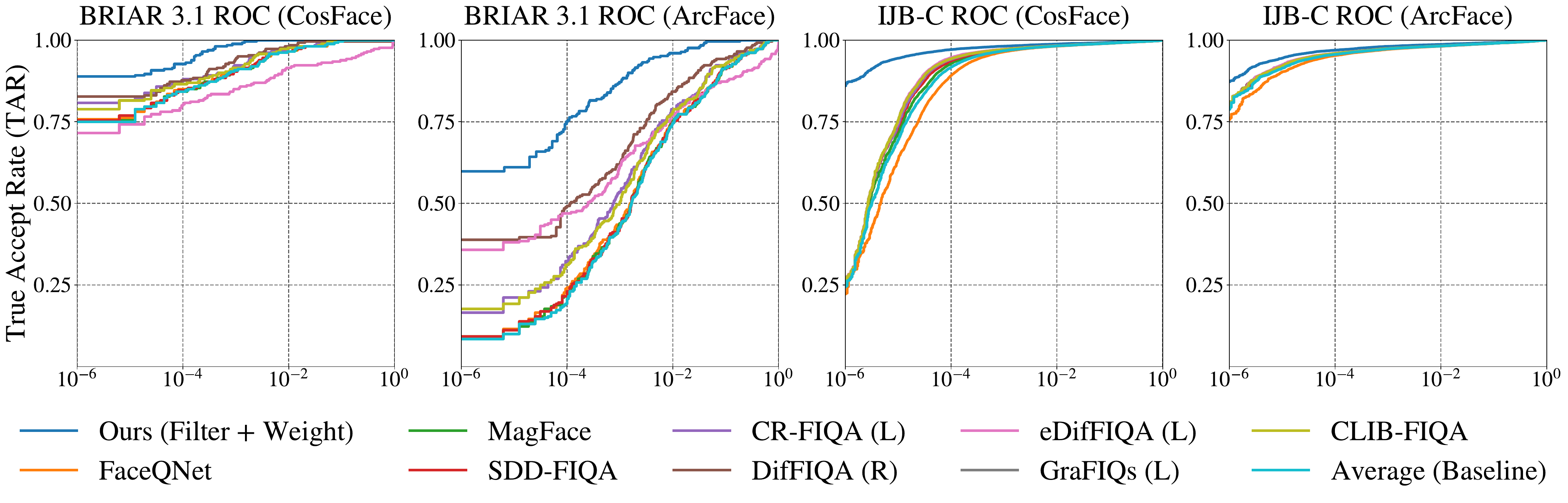}
  \includegraphics[width=\linewidth]{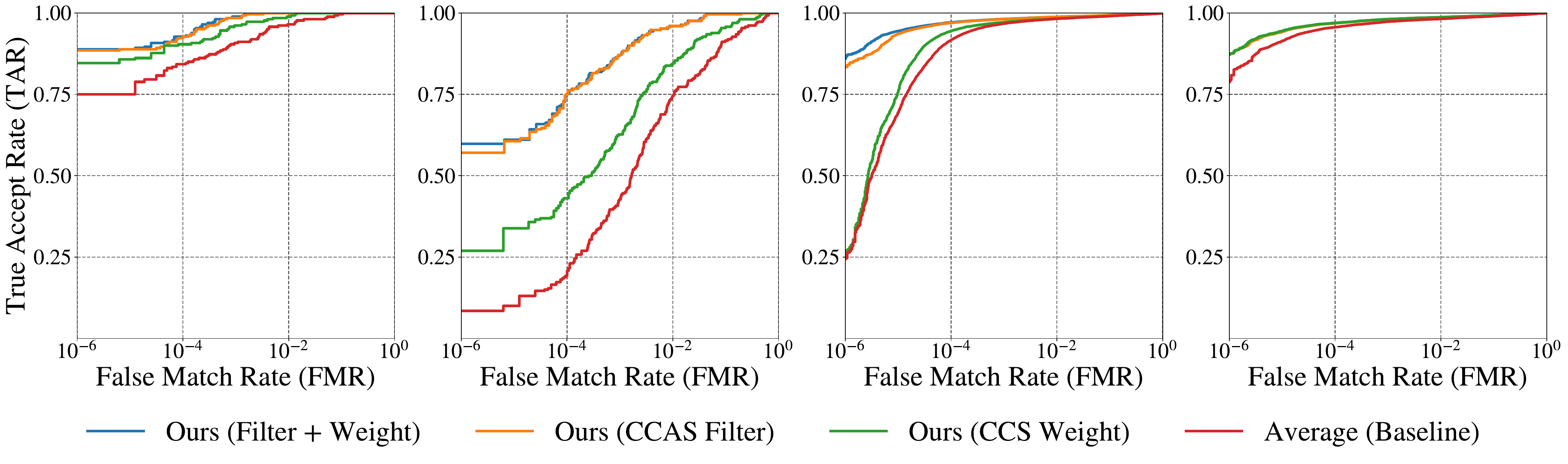}
  \vspace{-5mm}
  \caption{\textbf{Template-level ROC analysis on BRIAR Protocol~3.1~\cite{cornett2023expanding} and IJB-C~\cite{maze2018iarpa} with CosFace~\cite{wang2018cosface} and ArcFace~\cite{deng2019arcface}.}
  Top: strongest variant of each FIQA method.
  Bottom: individual and combined effects of CCAS filtering and CCS weighting.
  Curves correspond to Table~\ref{tab:tar_results}; \emph{Average} denotes uniform aggregation.}
  \label{fig:roc_all}
\end{figure*}

\begin{figure*}[t!]
  \centering
  \includegraphics[width=\linewidth]{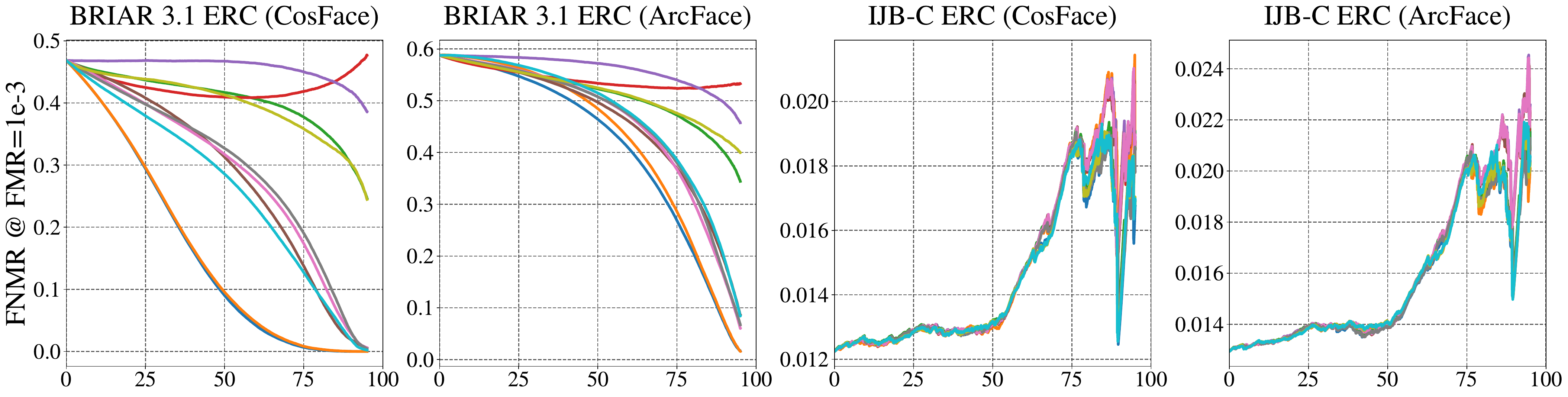}
  \includegraphics[width=\linewidth]{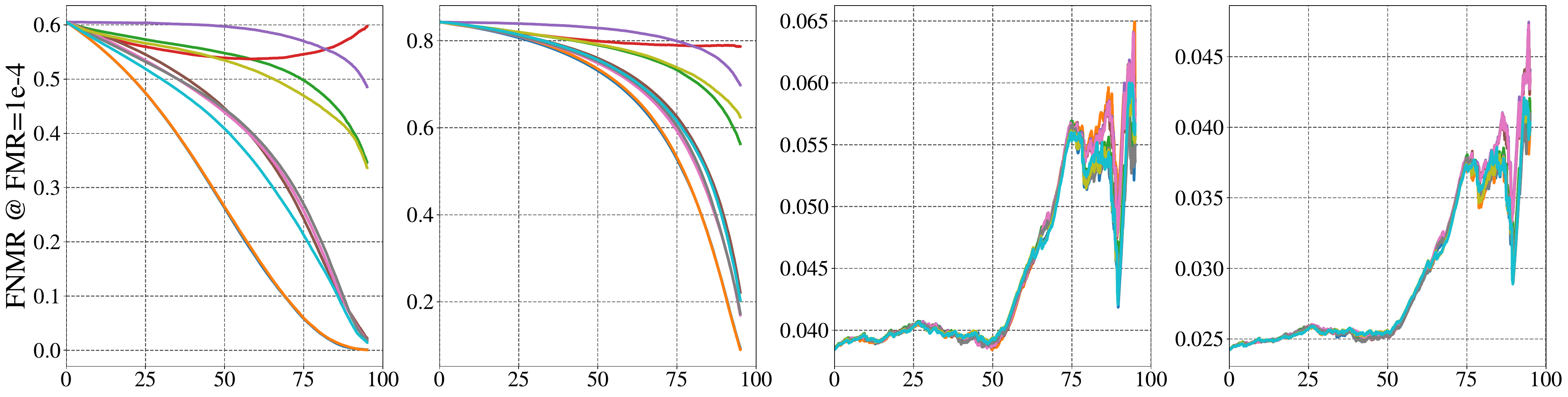}
  \includegraphics[width=\linewidth]{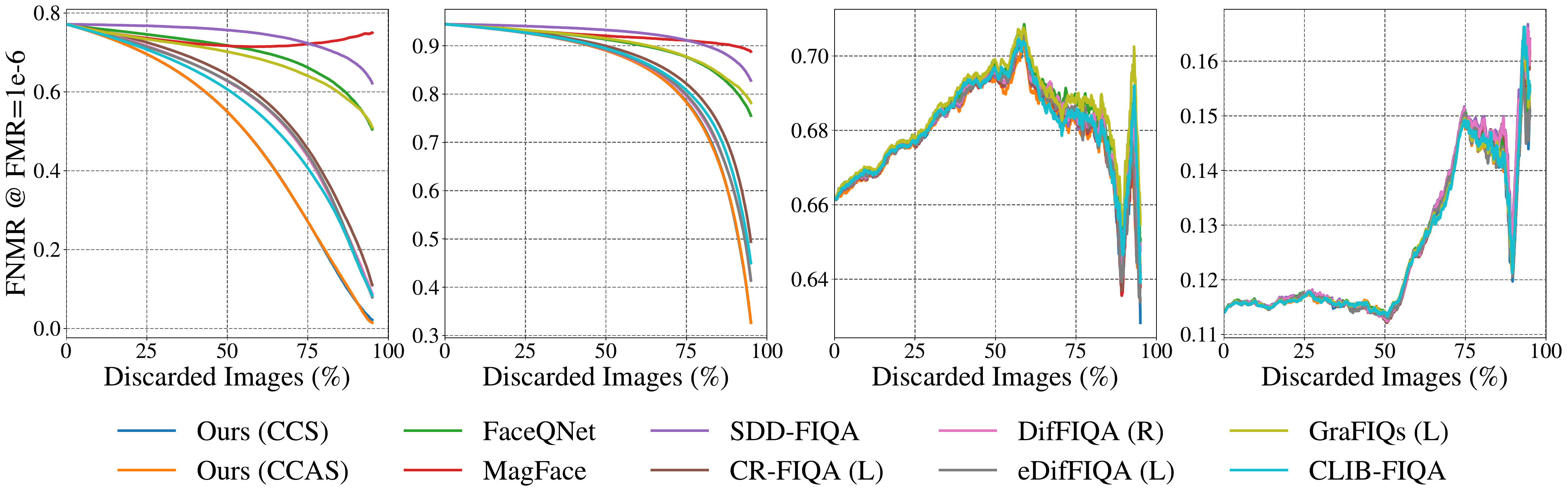}
  \vspace{-5mm}
  \caption{\textbf{Image-level ERC analysis on BRIAR Protocol~3.1~\cite{cornett2023expanding} and IJB-C~\cite{maze2018iarpa} with CosFace~\cite{wang2018cosface} and ArcFace~\cite{deng2019arcface}.}
  Curves closer to the bottom left indicate better trade-offs between discarding fraction and FNMR.
  Corresponding AUCs and Spearman correlations are reported in Table~\ref{tab:erc_auc_results}.}
  \label{fig:erc_all}
\end{figure*}

Figs.~\ref{fig:roc_all} and~\ref{fig:erc_all} expand the face verification and recognizability results of Sections~\ref{sec:experiments:aggregation} and~\ref{sec:experiments:erc}.
Fig.~\ref{fig:roc_all} shows that CCAS filtering produces the largest shift from uniform aggregation, especially at strict FMRs; CCS weighting adds smaller gains, and their combination yields the strongest overall curves.
Fig.~\ref{fig:erc_all} shows complementary image-level behavior: CCS and CCAS generally remain below competing ERC curves across discard fractions on BRIAR, so the gains are not tied to a single rejection threshold.
On IJB-C, differences narrow at higher FMRs, but TransFIRA remains among the strongest methods and the ordering is stable across operating points.

\subsection{Cross-Dataset Generalization}
\label{sec:experiments:crossdataset}

Domain-matched finetuning is typically strongest, pairing CosFace with BRIAR~\cite{wang2018cosface,cornett2023expanding} and ArcFace with WebFace for IJB-C~\cite{deng2019arcface,zhu2021webface260m,maze2018iarpa}.
\begin{table}[t]
\centering
\caption{\textbf{Ablation on training scope and predicted label.}\\Image-level results on BRIAR Protocol~3.1~\cite{cornett2023expanding} with CosFace~\cite{wang2018cosface}. \emph{P}, \emph{B}, and \emph{H} mark pretrained initialization, backbone finetuning, and head finetuning; \emph{Label} is the prediction target, where CR is CR-FIQA's Certainty Ratio~\cite{Boutros_2023_CVPR}. SC is Spearman correlation, and the best results are \textbf{bolded}.}
\begin{tabular}{cccc|cccc}
\toprule
\multicolumn{4}{c|}{\textbf{Method}} & \multicolumn{4}{c}{\textbf{BRIAR~\cite{cornett2023expanding}: FNMR-ERC AUC}} \\
P & B & H & Label & SC & $10^{-3}$ & $10^{-4}$ & $10^{-6}$ \\
\midrule
\checkmark &  & \checkmark & CCS & 0.4858 & 0.2422 & 0.4020 & 0.6278  \\
 & \checkmark & \checkmark & CCS & 0.7448 & 0.1791 & 0.2990 & 0.4994 \\
\checkmark & \checkmark & \checkmark & CR & 0.8553 & 0.1549 & 0.2739 & 0.4770 \\
\checkmark & \checkmark & \checkmark & CCAS & 0.8230 & 0.1558 & 0.2734 & \textbf{0.4752}  \\
\checkmark & \checkmark & \checkmark & CCS & \textbf{0.8566} & \textbf{0.1536} & \textbf{0.2722} & 0.4753 \\
\bottomrule
\end{tabular}
\label{tab:component_ablation}
\end{table}
\begin{table*}[t]
\centering
\caption{\textbf{Comparison with CR-FIQA filtering.}
TAR at fixed FMRs. CR-FIQA~\cite{Boutros_2023_CVPR} uses the midpoint cutoff $CR>0.5$, while TransFIRA uses the encoder-grounded cutoff $CCAS>0$. \emph{Filter} denotes filtering alone; \emph{Filter + Weight} combines filtering with the corresponding weighting score. The best result in each column is \textbf{\underline{bolded and underlined}}; the second best is \textbf{bolded}.}
\renewcommand{\arraystretch}{1.2}
\resizebox{\textwidth}{!}{%
    \begin{tabular}{l|ccc|ccc|ccc|ccc}
    \toprule
    \multirow{3}{*}{\textbf{Method}} & \multicolumn{6}{c|}{\textbf{BRIAR Protocol 3.1~\cite{cornett2023expanding}: TAR at Fixed FMR}} & \multicolumn{6}{c}{\textbf{IJB-C~\cite{maze2018iarpa}: TAR at Fixed FMR}} \\
     & \multicolumn{3}{c|}{\textbf{CosFace (BRIAR)~\cite{wang2018cosface, cornett2023expanding}}} & \multicolumn{3}{c|}{\textbf{ArcFace (WebFace)~\cite{guo2021insightface, zhu2021webface260m}}} & \multicolumn{3}{c|}{\textbf{CosFace (BRIAR)~\cite{wang2018cosface, cornett2023expanding}}} & \multicolumn{3}{c}{\textbf{ArcFace (WebFace)~\cite{guo2021insightface, zhu2021webface260m}}} \\
     & $10^{-3}$ & $10^{-4}$ & $10^{-6}$ & $10^{-3}$ & $10^{-4}$ & $10^{-6}$ & $10^{-3}$ & $10^{-4}$ & $10^{-6}$ & $10^{-3}$ & $10^{-4}$ & $10^{-6}$ \\
    \midrule
    CR-FIQA (L) (CR Filter) & 0.9077 & 0.8654 & 0.7808 & 0.4731 & 0.2577 & 0.1000 & 0.9782 & 0.9496 & 0.3617 & 0.9811 & 0.9686 & 0.8711 \\
    CR-FIQA (L) (CR Weight) & 0.9231 & 0.8731 & 0.8077 & 0.5346 & 0.3231 & 0.1654 & 0.9800 & 0.9617 & 0.3711 & 0.9811 & \textbf{0.9700} & \textbf{0.8726} \\
    CR-FIQA (L) (Filter + Weight) & 0.9231 & 0.8808 & 0.8231 & 0.5577 & 0.3385 & 0.1846 & 0.9800 & 0.9620 & 0.3712 & \textbf{0.9811} & \textbf{\underline{0.9701}} & \textbf{\underline{0.8726}} \\
    TransFIRA (CCAS Filter) & \textbf{0.9846} & \textbf{0.9231} & \textbf{0.8846} & \textbf{0.8715} & \textbf{0.7510} & \textbf{0.5703} & \textbf{0.9811} & \textbf{0.9697} & \textbf{0.8334} & \textbf{\underline{0.9811}} & 0.9685 & 0.8712 \\
    TransFIRA (CCS Weight) & 0.9615 & 0.9038 & 0.8462 & 0.6269 & 0.4308 & 0.2692 & 0.9718 & 0.9447 & 0.2512 & 0.9737 & 0.9578 & 0.7970 \\
    TransFIRA (Filter + Weight) & \textbf{\underline{0.9885}} & \textbf{\underline{0.9269}} & \textbf{\underline{0.8885}} & \textbf{\underline{0.8715}} & \textbf{\underline{0.7510}} & \textbf{\underline{0.5984}} & \textbf{\underline{0.9815}} & \textbf{\underline{0.9711}} & \textbf{\underline{0.8601}} & 0.9809 & 0.9693 & 0.8724 \\
    \bottomrule
    \end{tabular}%
}
\label{tab:filtering_comparison}
\end{table*}
\begin{table*}[t]
\centering
\caption{\textbf{Cross-dataset generalization results.} TAR at fixed FMRs when varying the recognition encoder, the initialization encoder, and the recognizability training data. Domain-matched configurations share the initialization and recognition encoder and use BRIAR training data for BRIAR and WebFace training data for IJB-C. The best result for each recognition encoder is \textbf{bolded}.}
\renewcommand{\arraystretch}{1.2}
\begin{tabular}{lll|*{3}{>{\centering\arraybackslash}p{1.45cm}}|*{3}{>{\centering\arraybackslash}p{1.45cm}}}
\toprule
\multirow{2}{*}{\shortstack[l]{\textbf{Recognition}\\\textbf{Encoder}}} & \multirow{2}{*}{\shortstack[l]{\textbf{Initialization}\\\textbf{Encoder}}} & \multirow{2}{*}{\shortstack[l]{\textbf{Training}\\\textbf{Data}}} & \multicolumn{3}{c|}{\textbf{BRIAR Protocol 3.1~\cite{cornett2023expanding}: TAR at Fixed FMR}} & \multicolumn{3}{c}{\textbf{IJB-C~\cite{maze2018iarpa}: TAR at Fixed FMR}} \\
 & & & $10^{-3}$ & $10^{-4}$ & $10^{-6}$ & $10^{-3}$ & $10^{-4}$ & $10^{-6}$ \\
\midrule
\multirow{4}{*}{CosFace~\cite{wang2018cosface}} & CosFace~\cite{wang2018cosface} & BRIAR~\cite{cornett2023expanding} & \textbf{0.9615} & \textbf{0.9038} & \textbf{0.8462} & \textbf{0.9720} & \textbf{0.9470} & 0.2493 \\
                         & ArcFace~\cite{guo2021insightface} & BRIAR~\cite{cornett2023expanding} & 0.9423 & 0.8923 & 0.8385 & 0.9716 & 0.9452 & 0.2255 \\
                         & CosFace~\cite{wang2018cosface} & WebFace~\cite{zhu2021webface260m} & 0.9423 & 0.8923 & 0.8192 & 0.9718 & 0.9447 & \textbf{0.2512} \\
                         & ArcFace~\cite{guo2021insightface} & WebFace~\cite{zhu2021webface260m} & 0.9000 & 0.8423 & 0.7500 & 0.9697 & 0.9276 & 0.2502 \\
\midrule
\multirow{4}{*}{ArcFace~\cite{guo2021insightface}} & CosFace~\cite{wang2018cosface} & BRIAR~\cite{cornett2023expanding} & 0.5846 & 0.3538 & 0.2115 & 0.9737 & 0.9594 & 0.7970 \\
                         & ArcFace~\cite{guo2021insightface} & BRIAR~\cite{cornett2023expanding} & \textbf{0.6269} & \textbf{0.4308} & \textbf{0.2692} & 0.9735 & 0.9593 & 0.7958 \\
                         & CosFace~\cite{wang2018cosface} & WebFace~\cite{zhu2021webface260m} & 0.5654 & 0.3423 & 0.1962 & 0.9738 & 0.9589 & 0.7978 \\
                         & ArcFace~\cite{guo2021insightface} & WebFace~\cite{zhu2021webface260m} & 0.4154 & 0.2154 & 0.0855 & \textbf{0.9809} & \textbf{0.9693} & \textbf{0.8724} \\
\bottomrule
\end{tabular}
\label{tab:cross-dataset}
\end{table*}
Deployment, however, rarely provides recognizability labels for each new distribution.
Table~\ref{tab:cross-dataset} extends Section~\ref{sec:experiments:aggregation} to mismatched training and evaluation domains, where TransFIRA remains highly competitive and sometimes matches or surpasses in-domain baselines.
Notably, CosFace finetuned on BRIAR achieves the best IJB-C TAR at $10^{-3}$ and $10^{-4}$ FMR.

Two properties likely support this transfer: the labels compare genuine and impostor identity centers, mirroring verification geometry, and CCAS encodes relative separation rather than absolute similarity.
The result spans a substantial shift from degraded BRIAR surveillance imagery to curated IJB-C web and video imagery.
Section~\ref{sec:video:transfer} extends this analysis to a harder setting by transferring a BRIAR-trained body predictor to unseen multi-camera video.

\begin{figure*}[t]
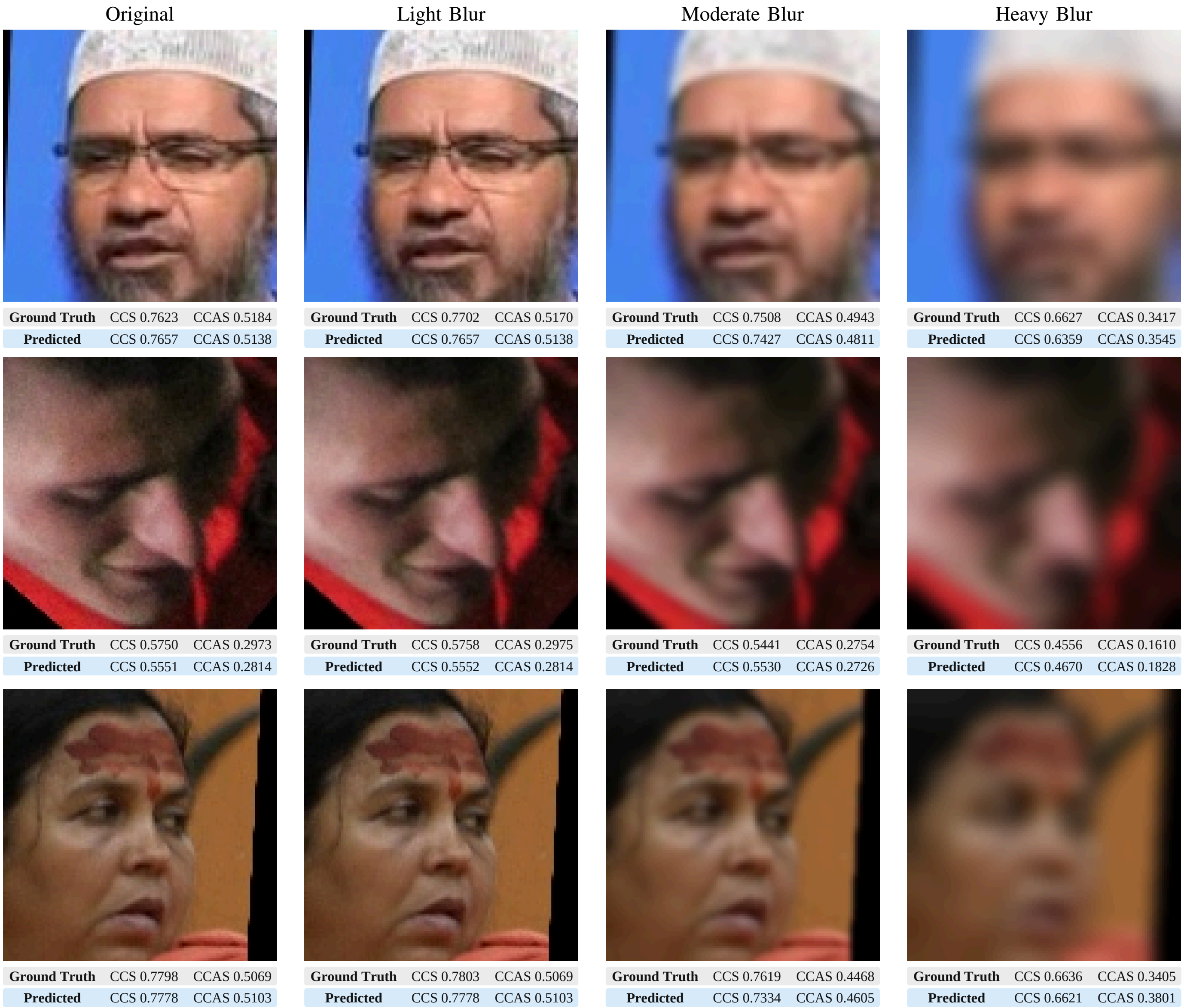

  \centering
  \setlength{\tabcolsep}{0pt}\small
  \begin{tabular}{@{}*{3}{>{\centering\arraybackslash}p{0.23235\linewidth}@{\hspace{0.02354\linewidth}}}>{\centering\arraybackslash}p{0.23235\linewidth}@{}}
  Original & Light Blur & Moderate Blur & Heavy Blur
  \end{tabular}\\[1pt]
  \includegraphics[width=\linewidth,trim=7.20 587.6 7.15 22.7,clip]{figures/explainability_selected.pdf}\\
  \includegraphics[width=\linewidth,trim=7.20 1.1 7.15 408.5,clip]{figures/explainability_selected.pdf}\\
  \caption{\textbf{Additional examples of recognizability under Gaussian blur with ArcFace~\cite{guo2021insightface,zhu2021webface260m} on IJB-C~\cite{maze2018iarpa}.}
  Light blur changes ground-truth CCS by less than $0.01$ in all examples and raises the top row from $0.762$ to $0.770$; predictions track the monotonic decline under stronger blur.}
  \label{fig:explainability_app}
\end{figure*}

\section{Video Identification Protocol Details}
\label{sec:appendix:video}

To reproduce the evaluation in Section~\ref{sec:video}, detections are assigned one-to-one to annotation boxes with $\mathrm{IoU}{\geq}0.5$, maximizing the match count first and total IoU second.
An annotated track enters the evaluation if it has at least five annotated frames, regardless of detection success, quality, or template availability.
Each face detection inherits the track of its associated body detection.
In \mevidvl{}, suffixed identity aliases are mapped to reviewed canonical identities and inherit the split of their base identity; one unresolved identity is excluded.
Identities are split by sorting a hash of each identifier within single- and multi-camera strata and assigning the first $30\%$ of each stratum to development.

Recognition embeddings are $L_2$-normalized before aggregation, and the aggregated template is normalized again before cosine scoring.
Frame weights are normalized over the frames that survive any filter, and a track with no surviving frames has no template.
Weighting does not change frame or template availability: constant scores under min--max scaling and all-zero weights under training-linear scaling fall back to uniform weights over the same frames.
Rank weights use ascending one-based score ranks within each track, with ties averaged.
The confidence filter in Table~\ref{tab:video_results} uses the development-set threshold whose retention best matches $CCAS>0$.

The fusion model is fitted on uniformly weighted development templates.
Face and body similarities are standardized by the mean and standard deviation of development nonmatch pairs; a missing modality contributes zero, and pairs with neither modality are unscored.
Coefficients are estimated by class-balanced logistic regression with $C=0.1$, remain unconstrained in sign, and are held fixed across conditions.

For each query, gallery tracks from the same camera are excluded, and the remaining tracks are ranked by fused score with exact ties broken deterministically.
Average precision includes every annotated cross-camera positive, even without a template; queries with no scored gallery items receive zero AP and \mbox{Rank-1}.
We obtain confidence intervals from $2{,}000$ query-identity bootstrap resamples shared across conditions and compute differences within each resample.
Galleries and fitted models remain fixed, and intervals are not adjusted for multiple comparisons.

For a budget of $K$ frames, body detections in each track are ordered by time, and those at positions $\lfloor (j+0.5)n/K \rfloor$, $j{=}0,\dots,K{-}1$, are retained before filtering; face detections are retained only with their associated body detections.
The transfer analysis in Section~\ref{sec:video:transfer} samples up to $64$ frames per track.
Identity centers are computed from all valid embeddings of the same split in cameras other than the probe's camera.
Each frame's empirical separation is its cosine similarity to the correct identity center minus its largest similarity to a competing identity center.
Spearman correlations are computed within each identity and averaged across identities, with $2{,}000$ identity-bootstrap resamples for the intervals.

\section{Additional Explainability Examples}
\label{sec:appendix:explainability}

Fig.~\ref{fig:explainability_app} extends the analysis of Section~\ref{sec:experiments:explainability} with three additional identities.
Light blur leaves CCS nearly unchanged in each case and slightly improves the top example, while moderate and heavy blur reduce recognizability along trajectories that the predictions follow.
These identity-specific trajectories complement the population averages in Table~\ref{tab:explainability} by revealing which samples remain stable under perturbation and which degrade rapidly.
They can flag difficult subjects, prompt re-enrollment or additional safeguards, and identify perturbations useful for robustness training.

\end{document}